\documentclass[runningheads]{llncs}
\usepackage{graphicx}
\usepackage{comment}
\usepackage{amsmath,amssymb} 
\usepackage{color}
\usepackage{cite}
\usepackage{subfig}
\usepackage{siunitx}

\usepackage{enumitem}

\usepackage{amssymb}
\usepackage{pifont}
\usepackage[dvipsnames]{xcolor}

\usepackage[width=122mm,left=12mm,paperwidth=146mm,height=193mm,top=12mm,paperheight=217mm]{geometry}

\usepackage[colorlinks, citecolor=green, linkcolor=black]{hyperref}

\begin{document}
\pagestyle{headings}
\mainmatter
\def\ECCVSubNumber{80}  

\title{AgingMapGAN (AMGAN): High-Resolution Controllable Face Aging with Spatially-Aware Conditional GANs}

\titlerunning{AgingMapGAN (AMGAN): High-Resolution Face Aging}
%
\author{Julien Despois, Fr\'ed\'eric Flament, Matthieu Perrot}
\authorrunning{Despois et al.}

\institute{L'Or\'eal Research \& Innovation}

\renewcommand{\floatpagefraction}{.7}%

\maketitle

\begin{abstract}
Existing approaches and datasets for face aging produce results skewed towards the mean, with individual variations and expression wrinkles often invisible or overlooked in favor of global patterns such as the fattening of the face. Moreover, they offer little to no control over the way the faces are aged and can difficultly be scaled to large images, thus preventing their usage in many real-world applications. To address these limitations, we present an approach to change the appearance of a high-resolution image using ethnicity-specific aging information and weak spatial supervision to guide the aging process. We demonstrate the advantage of our proposed method in terms of quality, control, and how it can be used on high-definition images while limiting the computational overhead.
\end{abstract}

\section{Introduction}

\begin{figure}
    \begin{minipage}{\textwidth}
    \centering
        \fbox{\includegraphics[width=11cm]{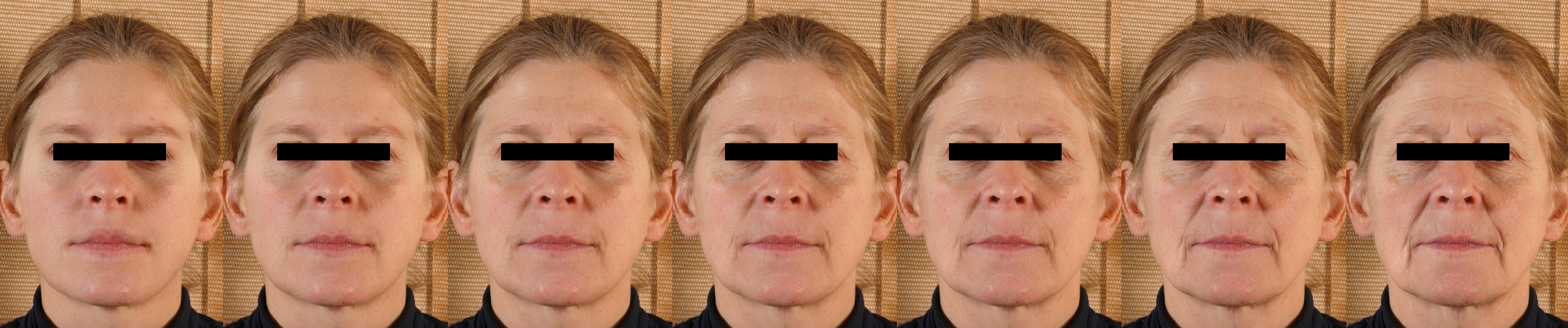}}
    \end{minipage}
    \begin{minipage}{\textwidth}
        \centering
        \includegraphics{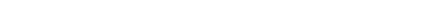}
   \end{minipage}
    \begin{minipage}{\textwidth}
        \centering
        \fbox{\includegraphics[width=11cm]{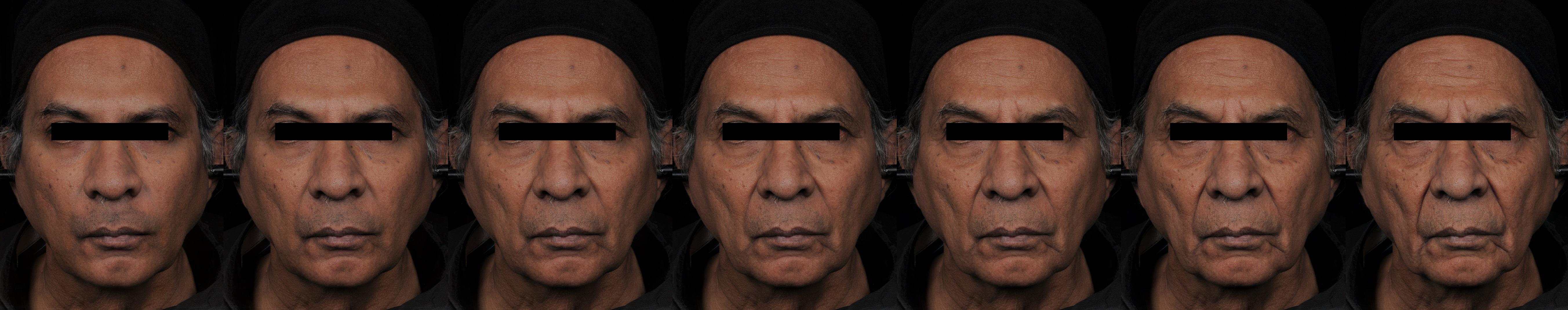}}
    \end{minipage}
\caption{High-resolution faces aged in a continuous manner with our approach}
\label{fig:sota_header}
\end{figure}

Face aging is an image synthesis task in which a reference image must be transformed to give the impression of a person of different age while preserving the identity and key facial features of the subject. When done correctly, this process can be used in various domains, from predicting the future appearance of a missing person to entertainment and educational uses. We focus on achieving high-resolution face aging, as it is a required step towards capturing the fine details of aging (fine lines, pigmentation, etc.). In recent years, Generative Adversarial Networks\cite{goodfellow2014generative} have allowed a learning-based approach for this task. The results, however, often lack in quality and only provide limited aging options. Popular models such as StarGAN\cite{choi2018stargan} fail to produce convincing results without additional tweaks and modifications. This partially stems from the choice of reducing aging to one's real or apparent age\cite{agustsson2017apparent}. Also, current approaches treat aging as a step-wise process, splitting age in bins (30-40, 40-50, 50+, etc.)\cite{zhu2019look, wang2018face, heljakka2018recursive, zeng2019controllable, antipov2017face}.

In reality, aging is a continuous process that can take many forms depending on genetic factors such as facial features and ethnicity, as well as lifestyle choices (smoking, hydration, sun damage, etc.) or behavior. Notably, expression wrinkles are promoted by habitual facial expressions and can be prominent on the forehead, upper lip, or at the corner of the eyes (crow's feet). In addition, aging is subjective as it depends on the cultural background of the person assessing the age. These factors call for a more fine-grained approach to face aging.

In this paper, we aim to obtain high-resolution face aging results by creating a model capable of individually transforming the local aging signs. Our contributions are as follows:

\begin{itemize}[leftmargin=1cm, itemsep=1mm]
  \item We show that a curated high-resolution dataset in association with a combination of novel and existing techniques produces detailed state-of-the-art aging results.
  \item We demonstrate how clinical aging signs and weak spatial supervision allows fine-grained control over the aging process of the different parts of the face.
  \item We introduce a patch-based approach to enable inference on high-resolution images while keeping the computational cost of training the model low.
\end{itemize}

\section{Related Work}

\subsubsection{Conditional Generative Adversarial Networks}
Generative Adversarial Networks\cite{goodfellow2014generative} leverage the principle of an adversarial loss to force samples generated by a generative model to be indistinguishable from real samples. This approach led to impressive results, especially in the domain of image generation. GANs can be extended to generate images based on one or several conditions. The resulting Conditional Generative Adversarial Networks are trained to generate images that satisfy both the realism and condition criteria.

\subsubsection{Unpaired Image-to-Image Translation}
Conditional GANs are a powerful tool for image-to-image translation\cite{isola2017image} tasks, where an input image is given to the model to synthesize a transformed image. StarGAN\cite{choi2018stargan} introduced a way to use an additional condition to specify the desired transformation to be applied. They propose to feed the input condition to the generator in the form of feature maps\cite{choi2018stargan} concatenated to the input image, but new approaches use more complex mechanisms such as AdaIN\cite{karras2019style} or its 2D extension SPADE\cite{park2019semantic} to give the generator the condition in a more optimal manner.
Where previous techniques required pixel-aligned training images in the different domains, recent works such as CycleGAN\cite{zhu2017unpaired} and StarGAN\cite{choi2018stargan} introduced a cycle-consistency loss to enable unpaired training between discrete domains. This has been extended in \cite{pumarola2018ganimation} to allow translation between continuous domains.

\subsubsection{Face Aging}
To age a face from a single picture, traditional approaches use training data of either one\cite{antipov2017face, zhu2018facial, heljakka2018recursive, zhu2019look, zeng2019controllable, upchurch2017deep} or multiple images\cite{wang2018face, song2018dual} of the same person, along with the age of the person when the picture was taken. The use of longitudinal data, with multiple photos of the same person, offers less flexibility as it creates a heavy time-dependent constraint on the dataset collection.

The age is usually binned into discrete age groups (20-30, 30-40, 40-50, 50+, etc.)\cite{antipov2017face, heljakka2018recursive, zhu2019look, zeng2019controllable}, which frames the problem more simply, but limits the control over the aging process and doesn't allow the training to leverage the ordered nature of the groups. \cite{zhu2018facial, upchurch2017deep} address this limitation by considering age as a continuous value. However, aging isn't objective because different skin types age differently, and different populations look for different signs of aging. Focusing on the apparent age as the guide for aging thus freezes the subjective point of view. Such approaches cannot be tailored to a population's perspective without requiring additional age estimation data from their point of view.

To improve the quality and level of details of the generated images, \cite{zhu2019look} use the attention mechanism from \cite{pumarola2018ganimation} in the generator. The generated samples are, however, low-definition images which are too coarse for real-world applications. Working at this scale hides some difficulties of generating realistic images, such as skin texture, fine lines, and the overall sharpness of the details. 

\section{Proposed Approach}
\subsection{Problem Formulation}
In this work, our goal is to use single unpaired images to train a model able to generate realistic high-definition ($1024 \times 1024$) aged faces, with continuous control over the fine-grained aging signs to create smooth transformations between the original and transformed images. This is a more intuitive approach, as aging is a continuous process and age group bins do not explicitly enforce a logical order.

We propose the use of ethnic-specific skin atlases\cite{bazin2007skin, bazin2010skin, bazin2012skin, bazin2015skin, flament2017skin} to incorporate the ethnic dimension of clinical aging signs. These atlases define numerous clinical signs such as the wrinkles underneath the eye, the ptosis of the lower part of the face, the density of pigmentary spots on the cheeks, etc. Each sign is linked to a specific zone on the face and scored on a scale that depends on ethnicity. Using these labels in addition to the age make for a more complete representation of aging, and allows transforming images with various combination of clinical signs and scores.

\begin{figure}
\begin{minipage}[b]{0.6\textwidth}
\centering
\subfloat[0.11]{\fbox{\includegraphics[height=1.3cm]{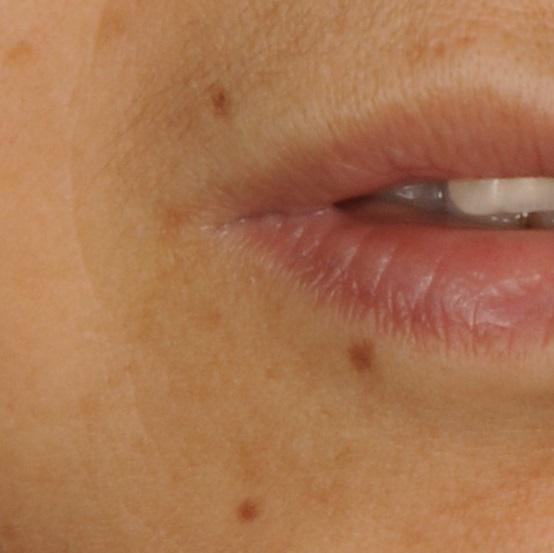}}}
\quad
\subfloat[0.36]{\fbox{\includegraphics[height=1.3cm]{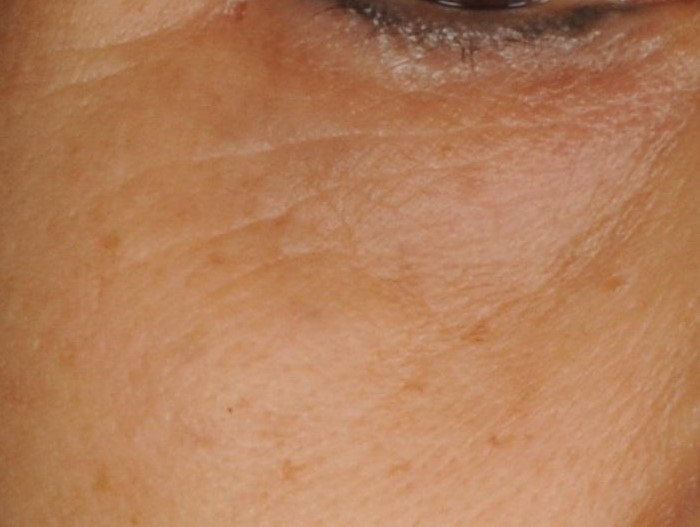}}}
\quad
\subfloat[0.31]{\fbox{\includegraphics[height=2cm]{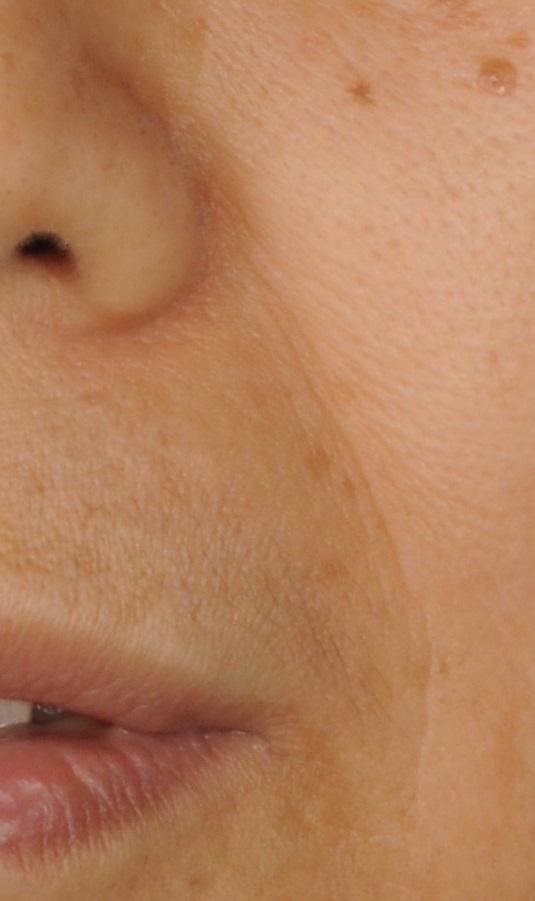}}}
\quad
\subfloat[0.40]{\fbox{\includegraphics[height=1.3cm]{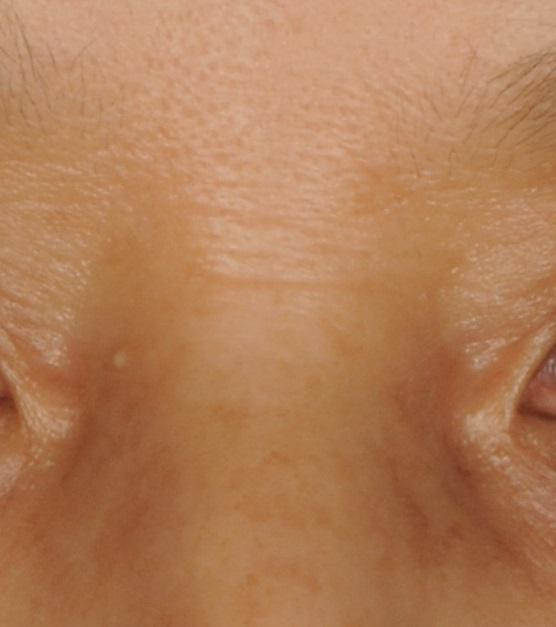}}}
\end{minipage}
\begin{minipage}[b]{0.4\textwidth}
\centering
    \subfloat[]{\includegraphics[height=2cm]{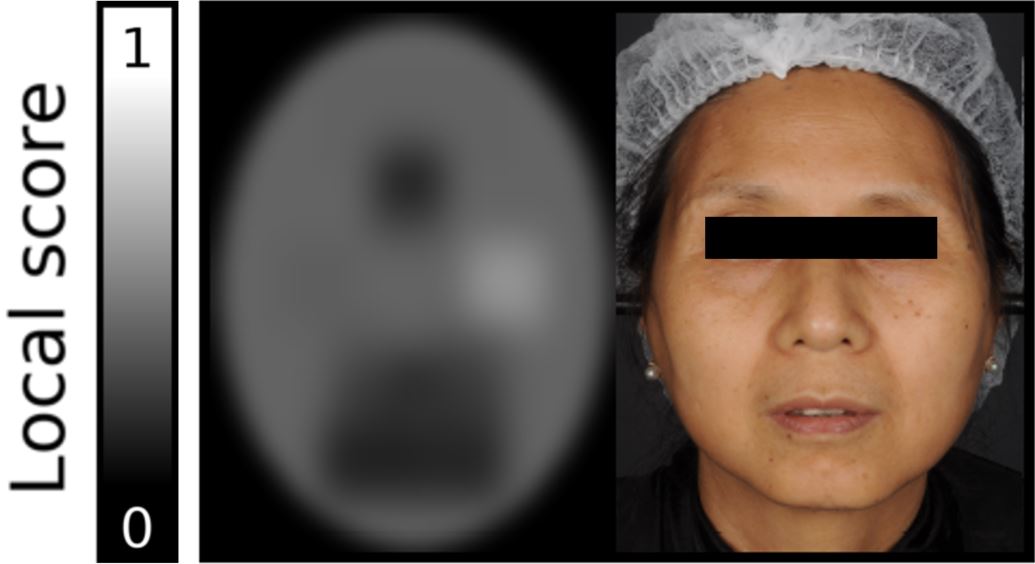}}
\end{minipage}
\caption{Aging sign zones (a-d) and their associated scores used to construct the aging map (e). The brightness of each pixel represents the normalized score of the localized clinical sign (wrinkles at the corner of the lips (a), underneath the eye wrinkles (b), nasolabial fold wrinkles (c), inter-ocular wrinkles (d), etc.) or the age where no sign is defined}
\label{fig:aging_maps}
\end{figure}

The aging target is passed to the network in the form of an aging map (Fig.~\ref{fig:aging_maps}). To do so, we compute facial landmarks and define the relevant zone for each aging sign. Each zone (e.g. forehead) is then filled with the score value of the corresponding sign (e.g. forehead wrinkles). We use the apparent age to fill in the blanks where the clinical signs are not defined. Finally, we coarsely mask the background of the image. 

Treating the whole image at once would be ideal, but training a model with $1024 \times 1024$ images requires large computational resources. Our approach allows us to train the model by patch, using only part of the image during training, and the corresponding part of the aging map. Patch-based training reduces the context (i.e. global information) for the task but also reduces the computational resources required to process high-resolution images in large batches, as recommended in \cite{brock2018large}. We leverage this to use a large batch size on small patches of $128 \times 128$, $256 \times 256$ or $512 \times 512$ pixels.

The major drawback of the patch-based training is that small patches can look similar (e.g. forehead and cheek) yet must be aged differently (e.g. respectively horizontal and vertical wrinkles). To avoid "mean" wrinkles on these ambiguous zones, we give the generator two patches coming respectively from a horizontal and a vertical gradient location map ({\fbox{\includegraphics[height=9pt]{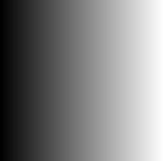}}}, {\fbox{\includegraphics[height=9pt]{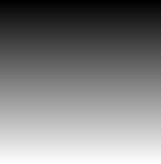}}}). This allows the model to know the position of the patch in order to differentiate between potentially ambiguous zones. 

\subsection{AMGAN - Network Architectures}
\begin{figure}
\centering
\includegraphics[width=12cm]{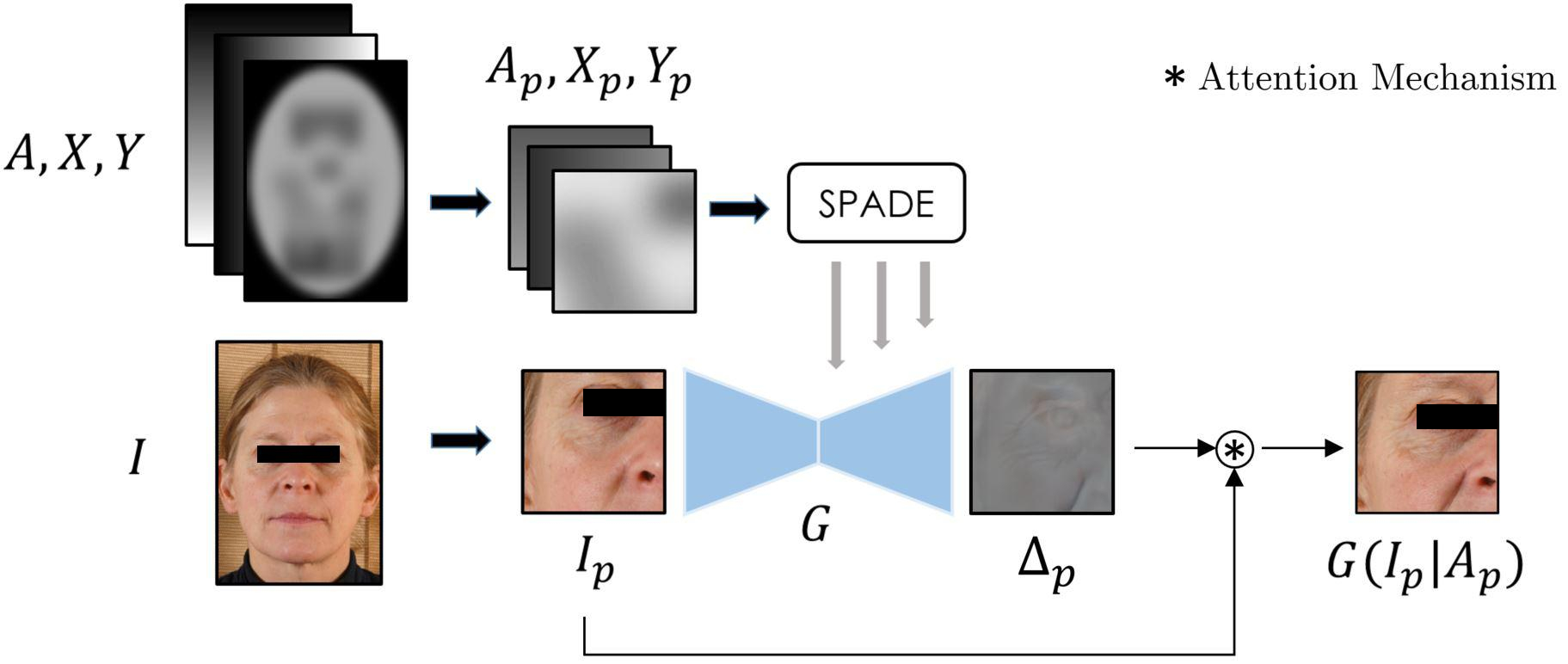}
\caption{Generator of our proposed patch-based training. We begin by cropping a patch from the image $I$, aging map $A$, and location maps $X$ and $Y$. The generator transforms the image patch $I_p$ according to the map and location}
\label{fig:model}
\end{figure}

\begin{figure}
\centering
\includegraphics[width=12cm]{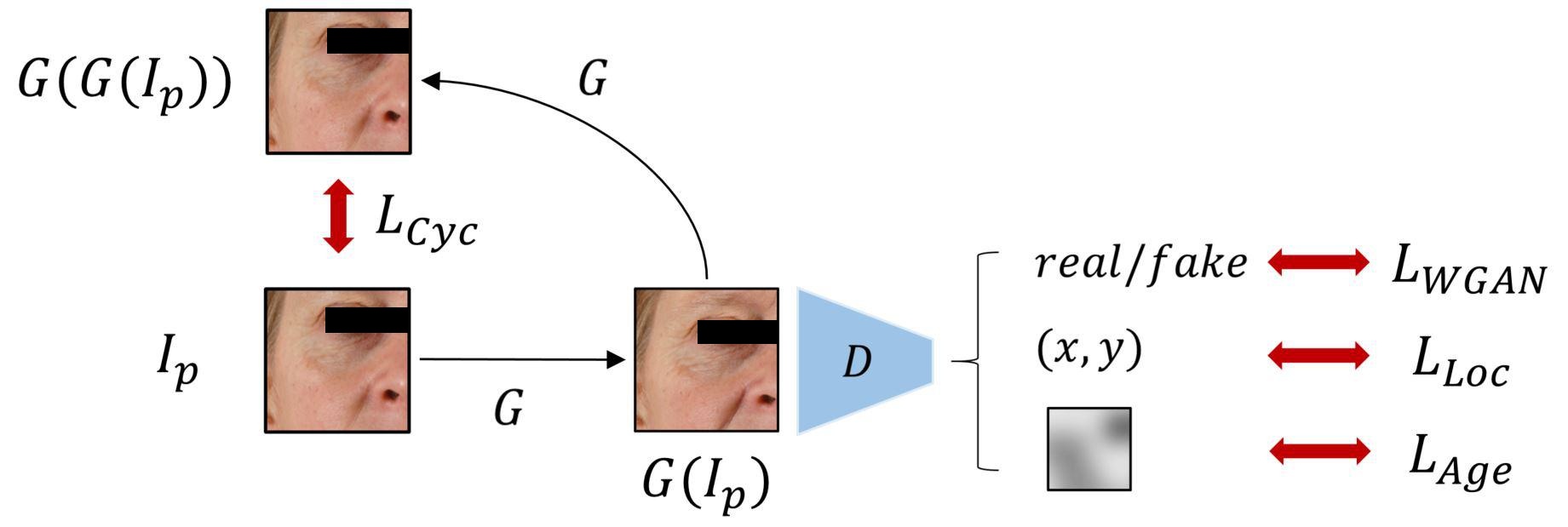}
\caption{The discriminator produces the real/fake output, the estimated location of the patch, and the estimated local aging map. The outputs are respectively penalized with the WGAN objective, location, and aging map loss functions. The cycle consistency loss ensures the transformation preserves the key features of the original image}
\label{fig:training}
\end{figure}

We base our training process on the StarGAN\cite{choi2018stargan} framework. Our generator is a fully convolutional encoder-decoder derived from \cite{choi2019stargan} with SPADE\cite{park2019semantic} residual blocks in the decoder to incorporate the aging and location maps. This allows the model to leverage the spatial information present in the aging map, and use it at multiple scales in the decoder. To avoid learning unnecessary details, we use the attention mechanism from \cite{pumarola2018ganimation} to force the generator to transform the image only where needed.
The discriminator is a modified version of \cite{choi2018stargan}, and produces the outputs for the WGAN\cite{arjovsky2017wasserstein} objective (given for an image $i$ and aging map $a$ in Equation~\ref{eq:loss_wgan}), the estimation of the coordinates of the patch, and the low-resolution estimation of the aging map. Fig.~\ref{fig:model} and Fig.~\ref{fig:training} present the patch-based training workflow.

\begin{align}\label{eq:loss_wgan}
  \mathcal{L}_{WGAN} = \mathbb{E}_{i}[D(i)] - \mathbb{E}_{i,a}[D(G(i,a))]\;
\end{align}

\subsection{Aging Maps}
To avoid penalizing the model for failing to place the bounding boxes with pixel-perfect precision, we blur the aging maps to smooth the edges and compute the discriminator regression loss on downsampled $10 \times 10$ maps. This formulation allows packing the information in a more compact and meaningful way than as individual uniform feature maps\cite{choi2018stargan, wang2018face, zhu2019look, zhu2018facial}. Our approach only requires multiple feature maps when there are large overlaps between signs (e.g. forehead pigmentation and forehead wrinkles). Considering an image patch $i$ and aging map patch $a$, the loss is given in Equation~\ref{eq:loss_age}.

\begin{align}\label{eq:loss_age}
  \mathcal{L}_{Age} = \mathbb{E}_{i}[\|a - D_{Age}(G(i,a))\|_{2}]\;
\end{align}

\subsection{Location Maps}
As in \cite{liu2018intriguing}, we use two orthogonal gradients as location maps ({\fbox{\includegraphics[height=9pt]{eccv2020kit/imgs/xy/x.jpg}}}, {\fbox{\includegraphics[height=9pt]{eccv2020kit/imgs/xy/y.jpg}}}) to help the generator apply relevant aging transformations to a given patch. The $(x,y)$ coordinates of the patch could be given to the generator as two numbers instead of linear gradients maps, but doing so would prevent the use of the model on the full-scale image as it would break its fully-convolutional nature. Considering an image patch $i$ and aging map patch $a$ located at coordinates $(x,y)$, the loss is given in Equation~\ref{eq:loss_loc}.

\begin{align}\label{eq:loss_loc}
  \mathcal{L}_{Loc} = \mathbb{E}_{i}[\|(x,y) - D_{Loc}(G(i,a))\|_{2}]\;
\end{align}

\subsection{Training}
The models are trained with the Adam\cite{kingma2014adam} optimizer with $\beta_1=0$, $\beta_2=0.99$ and learning rates of \SI{7e-5} for G and \SI{2e-4} for D. Following the two time-scale update rule\cite{heusel2017gans}, both models are updated at each step. Additionally, learning rates for both G and D are linearly decayed to zero over the course of the training. To enforce cycle-consistency, we use the perceptual loss\cite{zhang2018unreasonable} with $\lambda_{Cyc}=100$. For the regression tasks, we use $\lambda_{Loc}=50$ to predict the (x,y) coordinates of the patch and $\lambda_{Age}=100$ to estimate the downsampled aging map. The discriminator is penalized with the original gradient penalty presented in \cite{gulrajani2017improved} with $\lambda_{GP}=10$. Our complete loss objective function is given in Equation~\ref{eq:loss}.

\begin{align}\label{eq:loss}
  \mathcal{L} = \mathcal{L}_{WGAN} + \lambda_{Cyc}\mathcal{L}_{Cyc} + \lambda_{Age}\mathcal{L}_{Age} + \lambda_{Loc}\mathcal{L}_{Loc} + \lambda_{GP}\mathcal{L}_{GP} \;
\end{align}

\subsection{Inference}
 For inference, we use exponential moving average\cite{yazici2018unusual} over G's parameters. The trained generator can be used directly on the $1024 \times 1024$ image no matter the size of the patch used during training thanks to the fully convolutional nature of the network and the use of continuous 2D aging maps. We can either create a target aging map manually or use the face landmarks and target scores to build one.

\section{Experiments}

\subsection{Experimental Setting}
\begin{figure}
    \centering
    \begin{minipage}{0.3\textwidth}
        \hfill
    \end{minipage}
    \begin{minipage}{0.6\textwidth}
        \includegraphics[width=7.4cm]{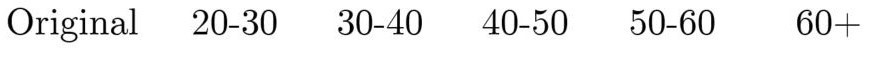}
    \end{minipage}

    \begin{minipage}{0.3\textwidth}
    (a) Wang et al.\cite{wang2018face}
    \end{minipage}
    \begin{minipage}{0.6\textwidth}
        \includegraphics[width=7.5cm]{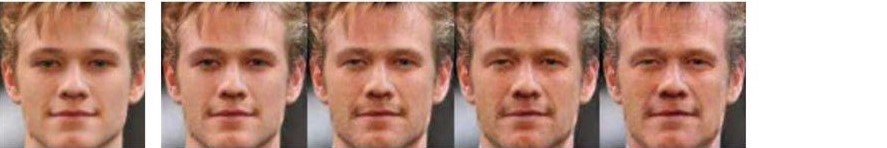}
    \end{minipage}
    
    \begin{minipage}{\textwidth}
        \includegraphics[height=01mm]{eccv2020kit/imgs/other_papers/small_spacer.jpg}
    \end{minipage}

    \begin{minipage}{0.3\textwidth}
    (b) Heljakka et al.\cite{heljakka2018recursive}
    \end{minipage}
    \begin{minipage}{0.6\textwidth}
        \includegraphics[width=7.5cm]{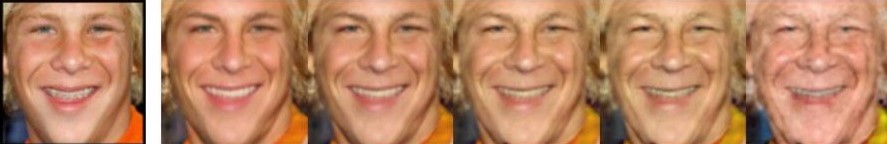}
    \end{minipage}

    \begin{minipage}{\textwidth}
        \includegraphics[height=1mm]{eccv2020kit/imgs/other_papers/small_spacer.jpg}
    \end{minipage}
    
    \begin{minipage}{0.3\textwidth}
    (c) Song et al.\cite{song2018dual}
    \end{minipage}
    \begin{minipage}{0.6\textwidth}
        \includegraphics[width=7.5cm]{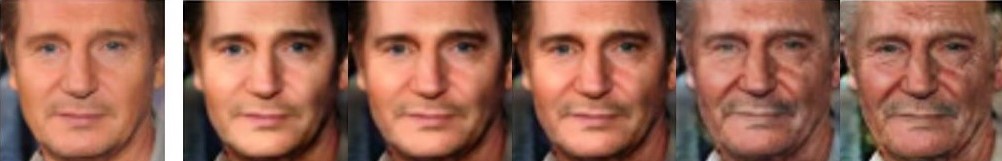}
    \end{minipage}

    \begin{minipage}{\textwidth}
        \includegraphics[height=1mm]{eccv2020kit/imgs/other_papers/small_spacer.jpg}
    \end{minipage}
    
    \begin{minipage}{0.3\textwidth}
    (d) Antipov et al.\cite{antipov2017face}
    \end{minipage}
    \begin{minipage}{0.6\textwidth}
        \includegraphics[width=7.5cm]{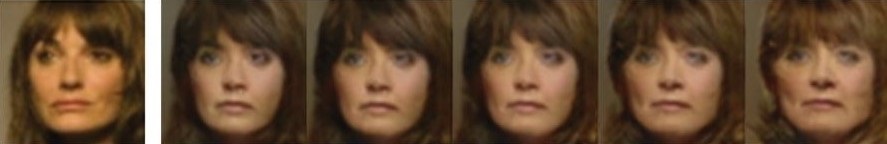}
    \end{minipage}
    
    \begin{minipage}{\textwidth}
        \includegraphics[height=1mm]{eccv2020kit/imgs/other_papers/small_spacer.jpg}
    \end{minipage}
    
    \begin{minipage}{0.3\textwidth}
    (e) Upchurch et al.\cite{upchurch2017deep}
    \end{minipage}
    \begin{minipage}{0.6\textwidth}
        \includegraphics[width=7.5cm]{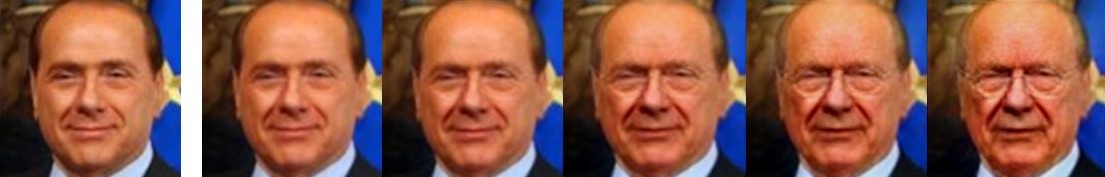}
    \end{minipage}
    
    \begin{minipage}{\textwidth}
        \includegraphics[height=1mm]{eccv2020kit/imgs/other_papers/small_spacer.jpg}
    \end{minipage}

    \caption{Previous approaches operate on low-resolution images and suffer from a lack of wrinkles dynamic range, especially for expression wrinkles (a). They are also prone to color shifts and artifacts (b, c, d, e), as well as unwanted correlated features such as the fattening of the face (d), or the addition of glasses (e)}
    \label{fig:other_works}
\end{figure}

Most face aging datasets\cite{rothe2015dex, chen2014cross, ricanek2006morph} suffer from a lack of diversity in terms of ethnicity\cite{karkkainen2019fairface}, and focus on low-resolution images (up to $250 \times 250$ pixels). This isn't sufficient to capture fine details related to skin aging. Moreover, they often fail to normalize the pose and expression of the faces (smiling, frowning, raised eyebrows), which results in accentuated wrinkles unrelated to aging (mostly nasolabial wrinkles, crow's feet wrinkles, forehead wrinkles and wrinkles underneath the eye). Finally, the lack of fine-grained information on the aging signs causes other approaches to capture unwanted correlated features such as the fattening of the face, as observed in datasets such as IMDB-Wiki\cite{rothe2015dex}. These effects can be observed in Fig.~\ref{fig:other_works}.

To address these issues, we tested our models on two curated high-resolution datasets, using generated aging maps or uniform aging maps to highlight the rejuvenation/aging.

\subsection{Flicker Faces High-Quality Dataset (FFHQ)}
To tackle the high-resolution aging problem, we have tested our approach on the FFHQ dataset \cite{karras2019style}. To minimize the issues in lighting, pose, and facial expressions, we applied simple heuristics to select a subset of the dataset of better quality. To do so, we extracted facial landmarks from all faces and used them to remove all images where the head was too heavily tilted left, right, up, or down. In addition, we removed all images with an open mouth to limit artificial nasolabial fold and underneath the eye wrinkles. Finally, we used a HOG\cite{dalal2005histograms} feature descriptor to remove images with hair covering the face. This selection brings down the dataset from 70k+ to 10k+ images. Due to the extreme diversity of the FFHQ dataset, the remaining images are still far from being perfect, especially in terms of lighting color, direction, and exposure.

To obtain the scores of the individual aging signs on these images, we used aging sign estimation models based on the ResNet\cite{szegedy2017inception} architecture that we trained on the dataset described in Section~\ref{section:dataset}. Finally, we generated the ground truth aging maps using the landmarks as a basis for the coarse bounding-boxes. We trained our model on $256 \times 256$ patches randomly selected on the $1024 \times 1024$ face.

\subsection{High-Quality Standardized Dataset}
\label{section:dataset}
To obtain better performance, we have collected a dataset of 6000 high-resolution ($3000 \times 3000$) images of faces, centered and aligned, spanning most ages (18-80), genders, and ethnicities (African, Caucasian, Chinese, Japanese and Indian). The images were labeled using ethnicity-specific clinical aging sign atlases\cite{bazin2007skin, bazin2010skin, bazin2012skin, bazin2015skin, flament2017skin} and scored on signs covering most of the face (apparent age, forehead wrinkles, nasolabial fold, underneath the eye wrinkles, upper lip wrinkles, wrinkles at the corner of the lips and ptosis of the lower part of the face).

\section{Results}
\subsection{FFHQ}
Despite the complexity of the dataset, and without ground truth age values, our patch-based model is able to transform the individual wrinkles on the face in a continuous manner. 

\begin{figure}
\centering
\fbox{\includegraphics[width=5.8cm]{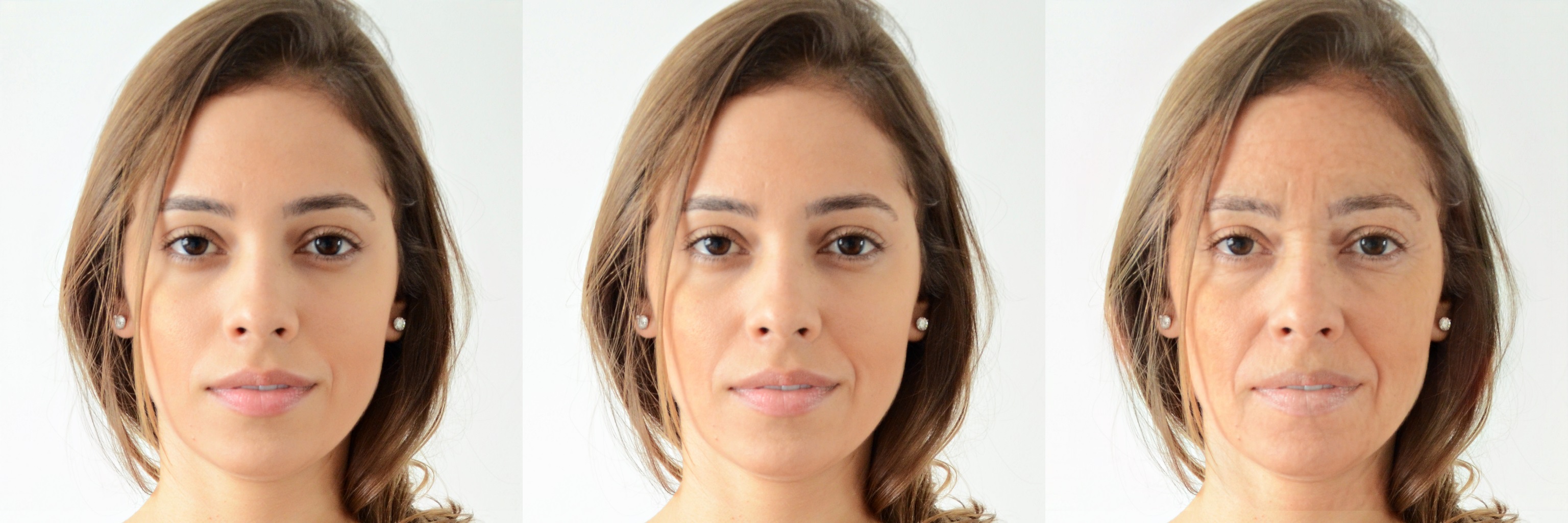}}
\centering
\fbox{\includegraphics[width=5.8cm]{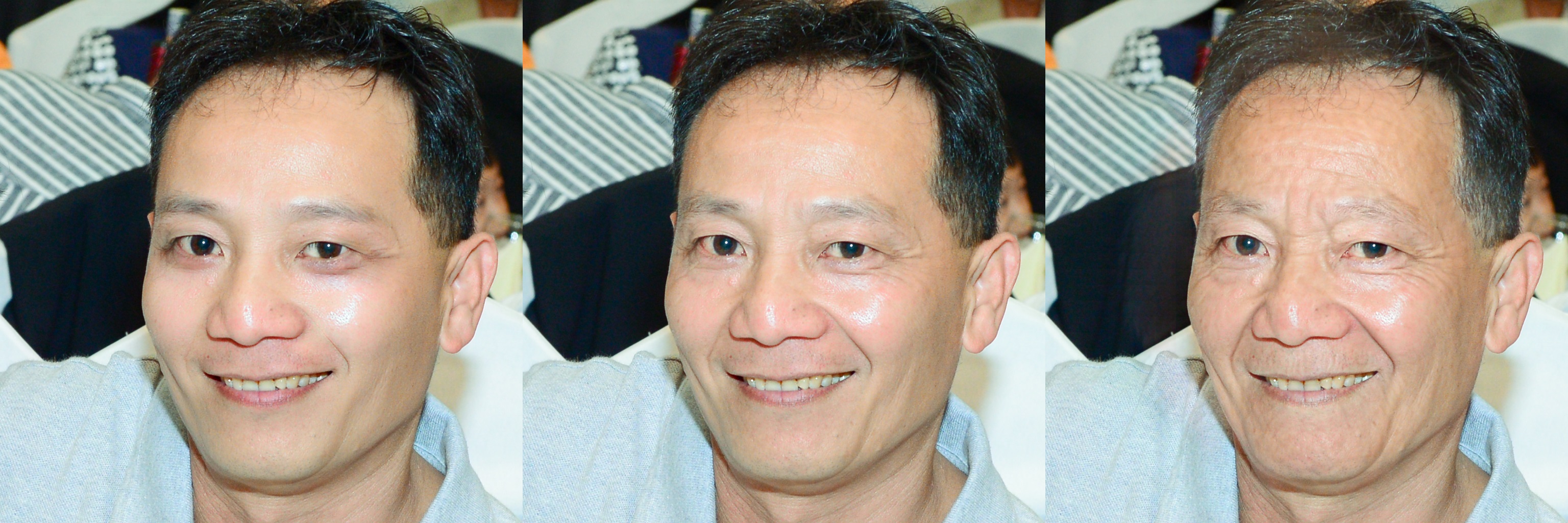}}
\centering
\fbox{\includegraphics[width=5.8cm]{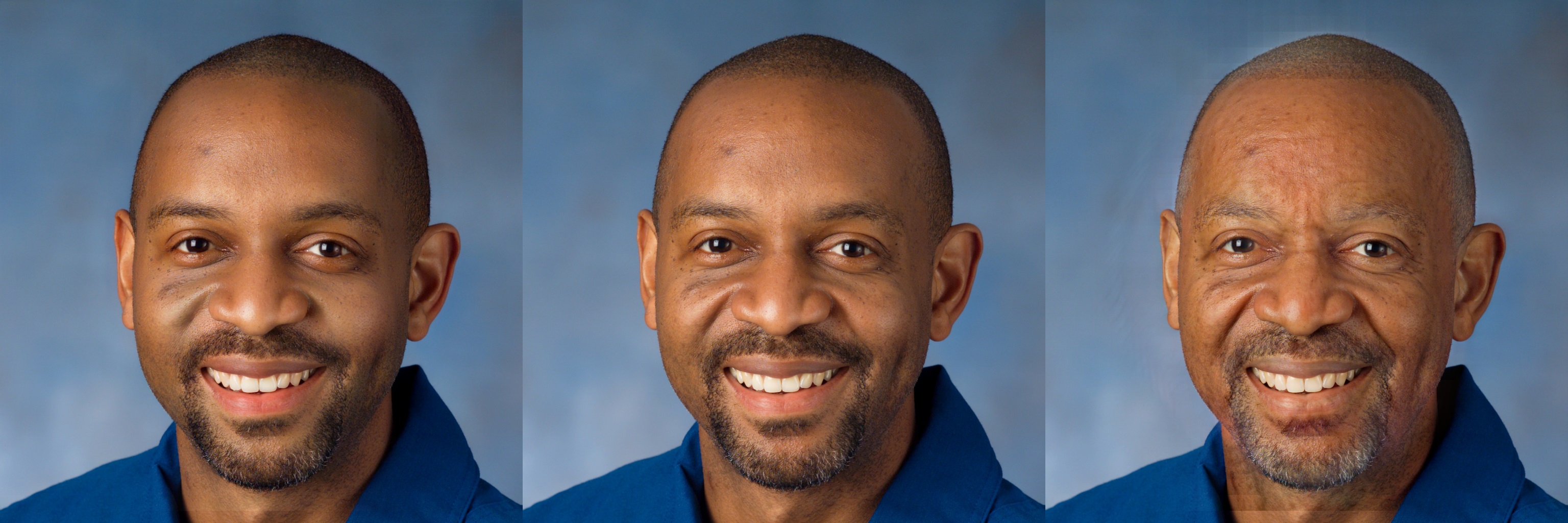}}
\centering
\fbox{\includegraphics[width=5.8cm]{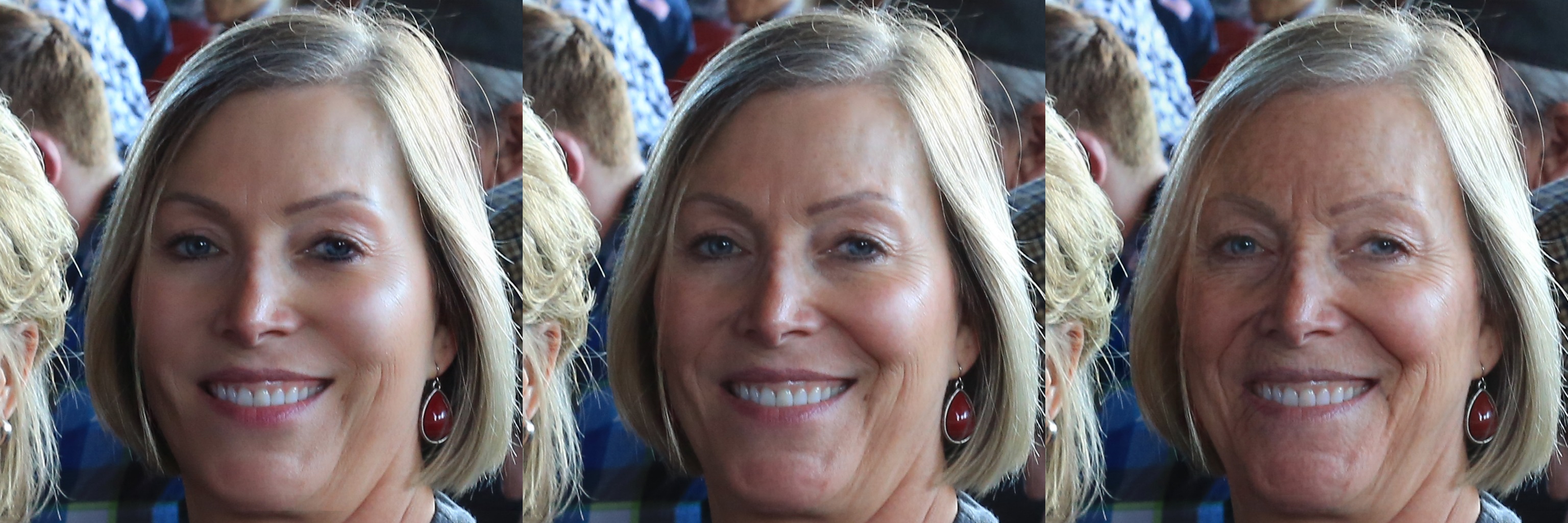}}
\centering
\fbox{\includegraphics[width=5.8cm]{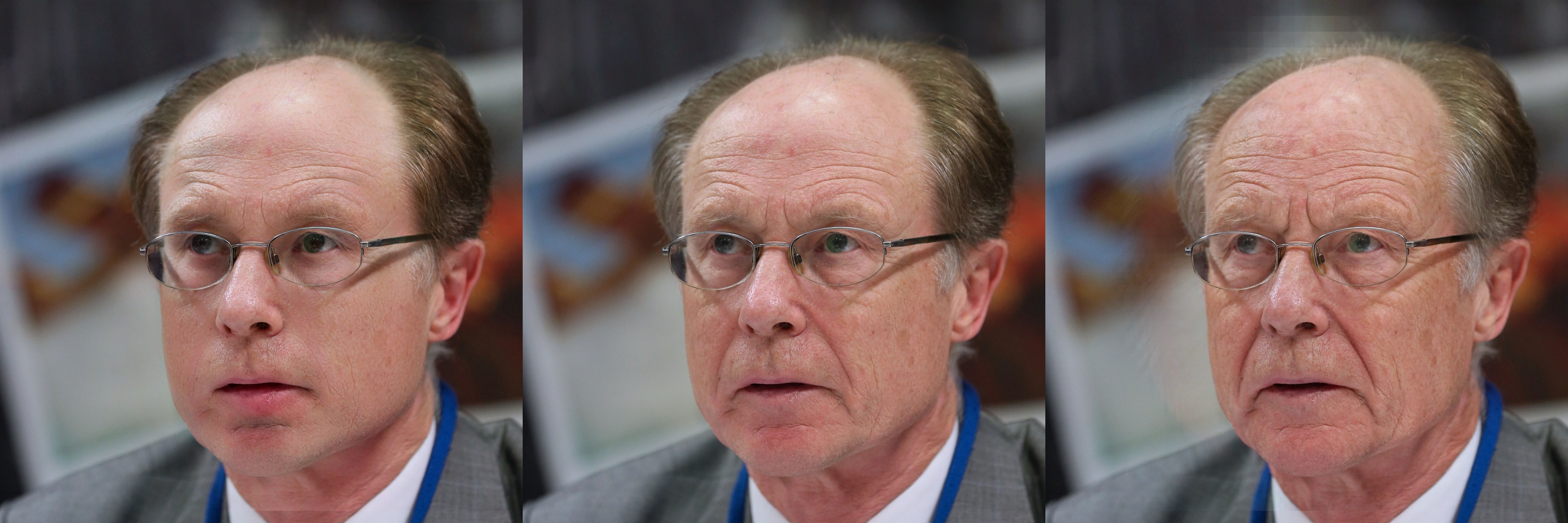}}
\centering
\fbox{\includegraphics[width=5.8cm]{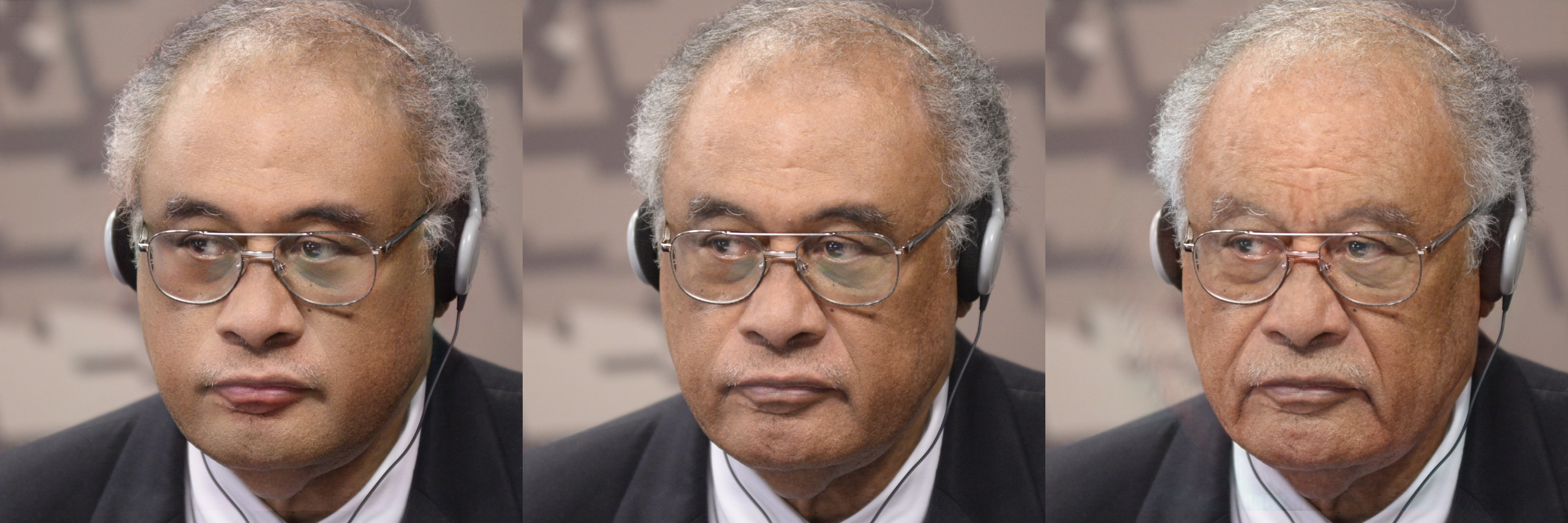}}

\caption{Rejuvenation ({\it left}), original ({\it center}) and aging ({\it right}) for faces of different age and ethnicity from FFHQ dataset using our approach}
\label{fig:ffhq}
\end{figure}

\begin{figure}
\centering
\fbox{\includegraphics[width=5.8cm]{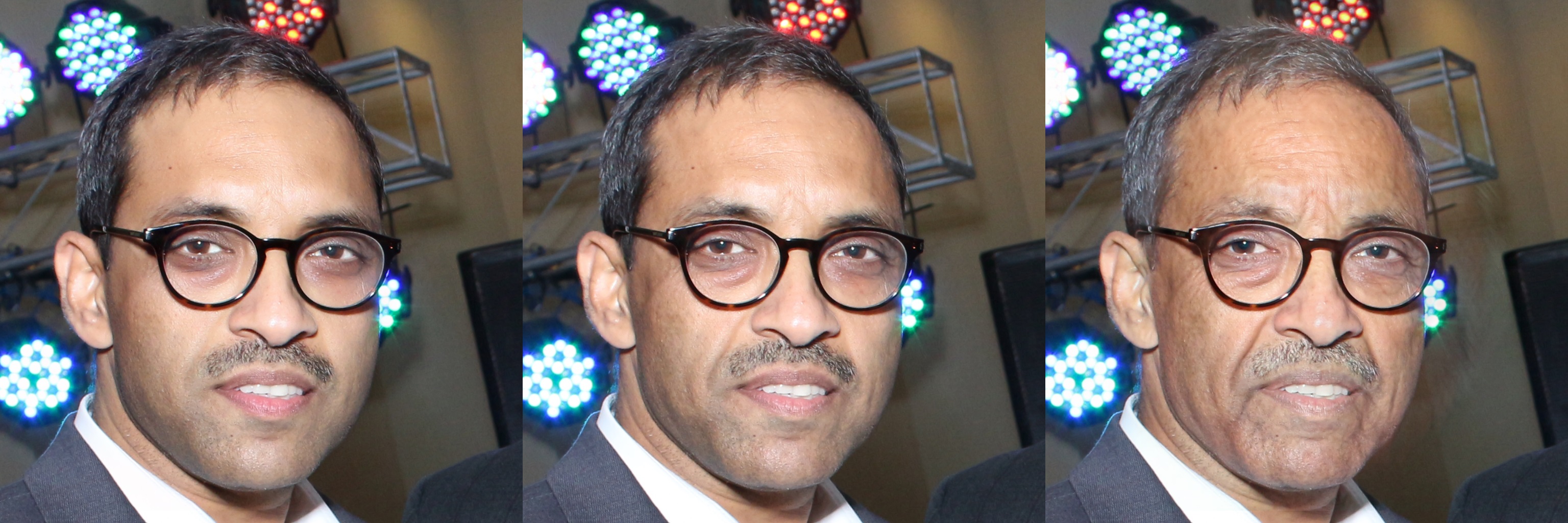}}
\centering
\fbox{\includegraphics[width=5.8cm]{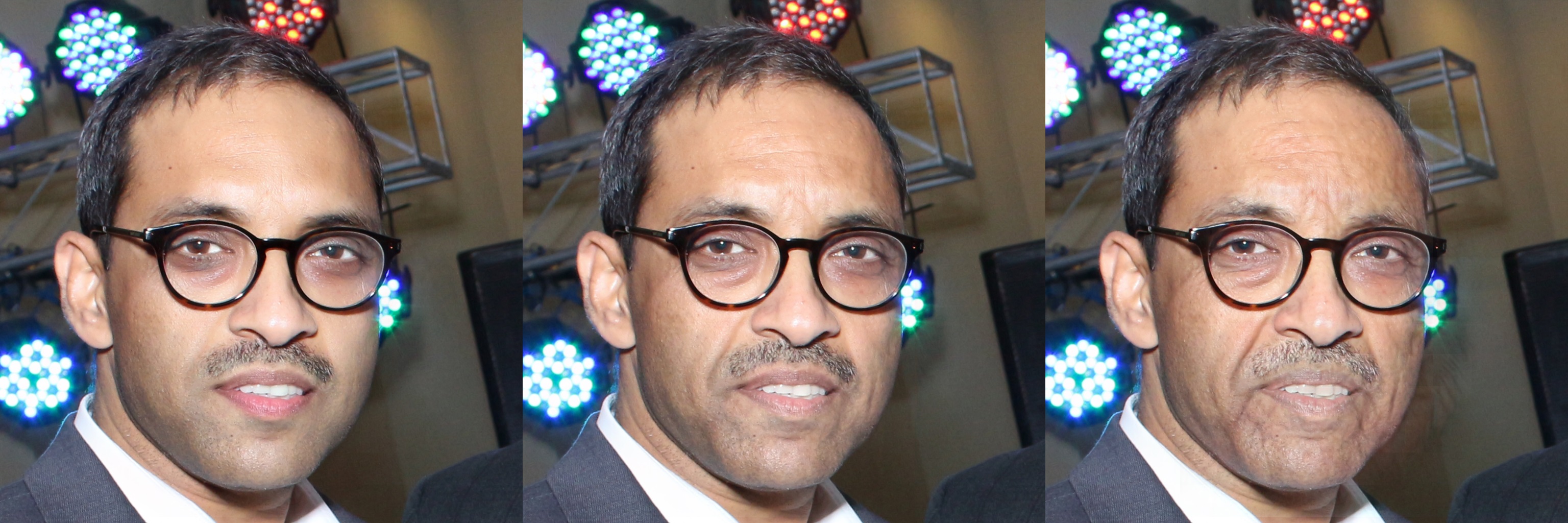}}
\centering
\caption{Where no sign is defined, we fill the map with the age. This helps the model learn global features like the greying of the hair ({\it left}). Using individual clinical signs in an aging map allows us to age all signs but keep the appearance of the hair intact ({\it right})}
\label{fig:ffhq_misc}
\end{figure}

Fig.~\ref{fig:ffhq} displays how the model was able to transform the different wrinkles despite the complexity of the patch-based training, the large variation in lighting in the dataset, and the unbalance between grades of clinical signs/age, with a vast majority of young subjects with few wrinkles. Fig.~\ref{fig:ffhq_misc} highlights the control we have over the individual signs, allowing aging the face in a controllable way that wouldn't be possible with the only label of the age.

\subsection{High-Quality Standardized Dataset}
\begin{figure}
\centering
\fbox{\includegraphics[width=11cm]{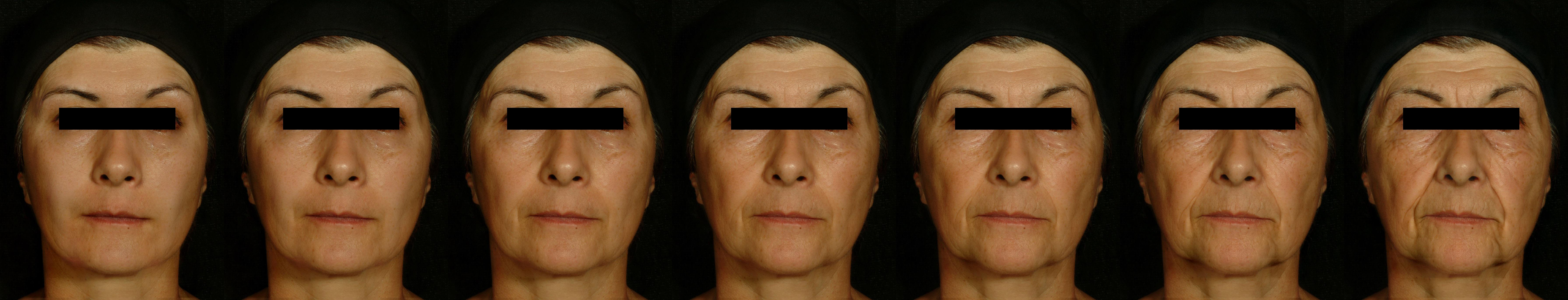}}
\centering
\fbox{\includegraphics[width=11cm]{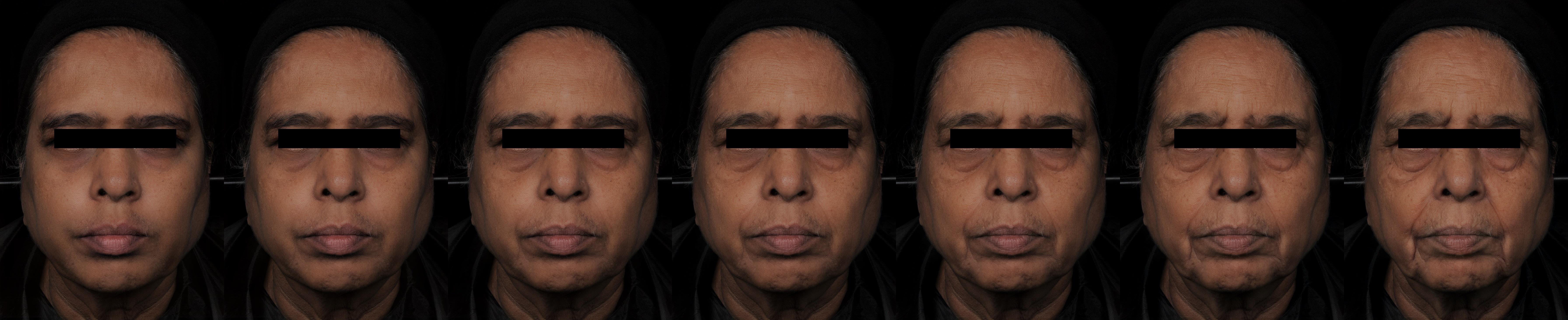}}
\centering
\fbox{\includegraphics[width=11cm]{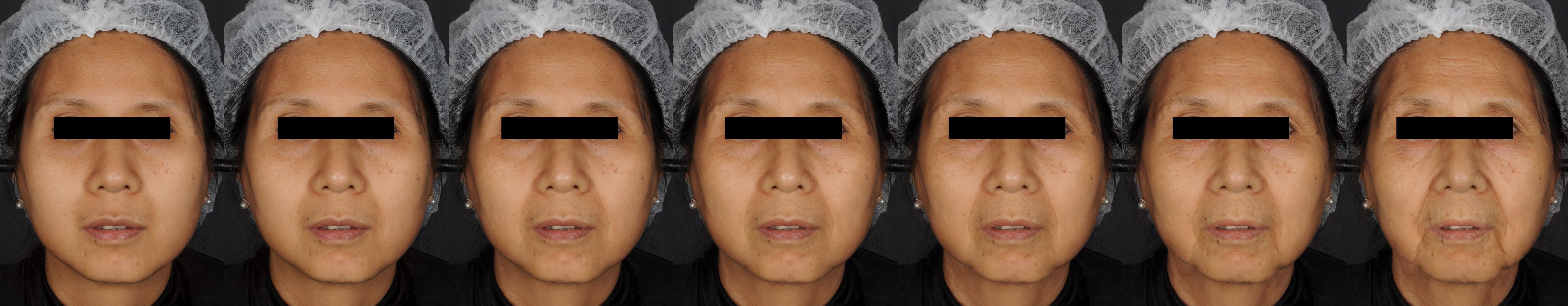}}
\centering
\fbox{\includegraphics[width=11cm]{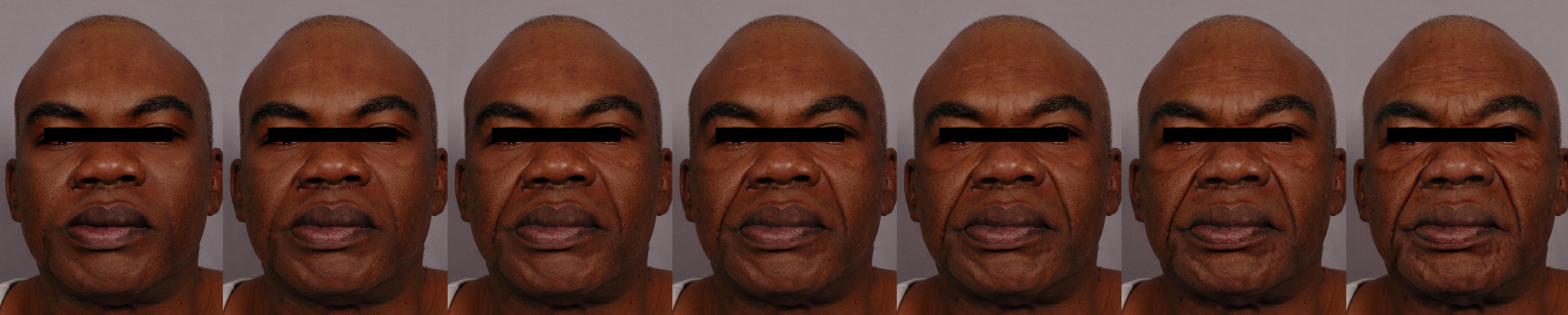}}
\caption{Faces aged in a continuous manner. No zone is left unchanged, even the forehead or the sagging of lower part of the face. The complementary age information used to fill the gap can be seen on the thinning or greying of the eyebrows. Note: zooming is recommended to see the fine details of the figure}
\label{fig:sota_full}
\end{figure}

On our standardized images, and with better coverage across ethnicity and aging signs, our model demonstrates state-of-the-art performance (Fig.~\ref{fig:sota_header}, Fig.~\ref{fig:sota_full}), with a high level of detail, realism, and no visible artifacts, color shifts or unwanted correlated features as seen in previous works (Fig.~\ref{fig:other_works}). The aging process is successful along the continuous spectrum of age maps, allowing realistic images to be generated for a diverse set of sign severity values (Fig.~\ref{fig:misc}). More examples as well as HD videos are available in the supplementary materials and at \url{https://despoisj.github.io/AgingMapGAN/}.

\begin{figure}
\centering
\subfloat[]{\fbox{\includegraphics[height=2.5cm]{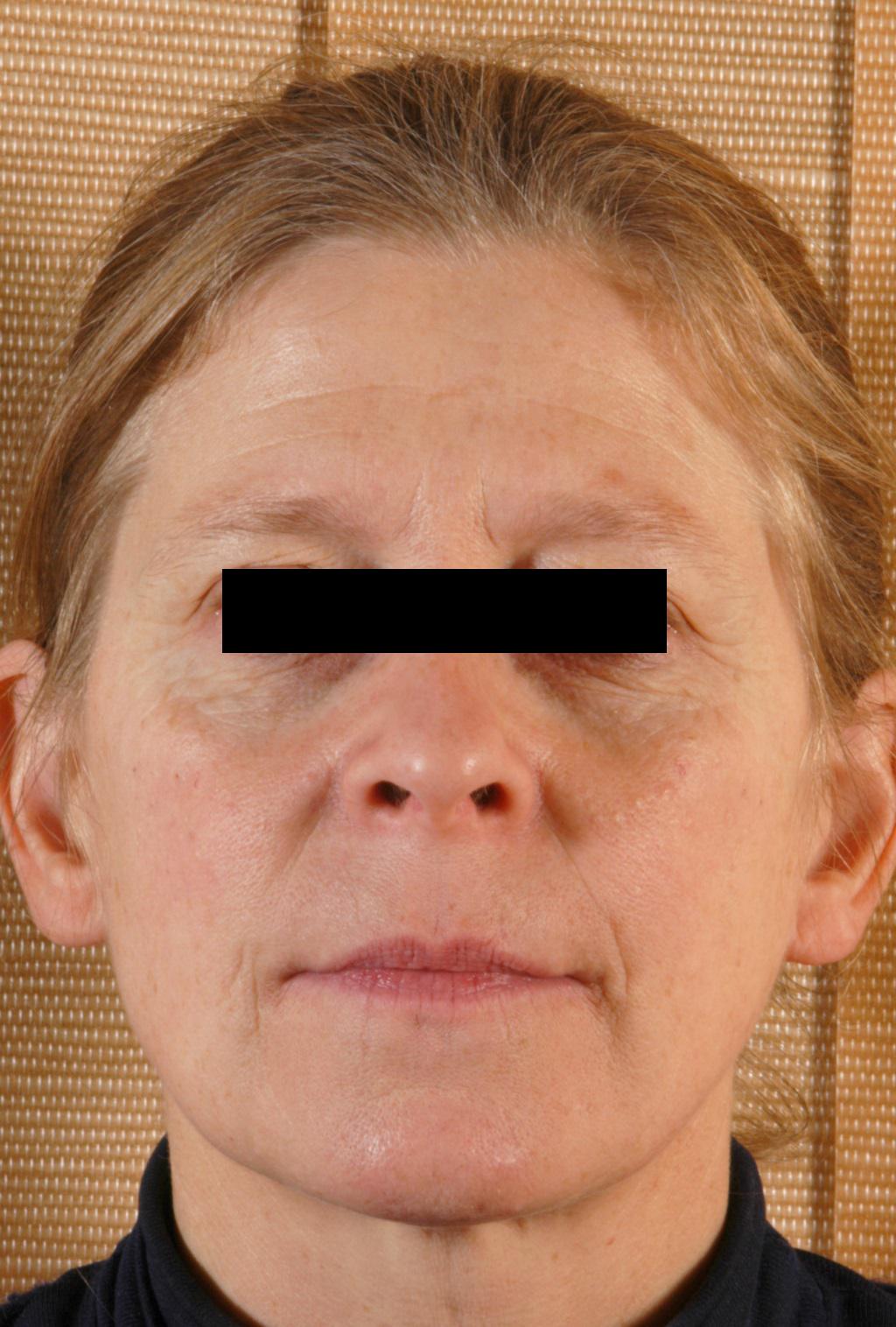}}\label{misc_original}}
\qquad
\subfloat[]{\fbox{\includegraphics[height=2.5cm]{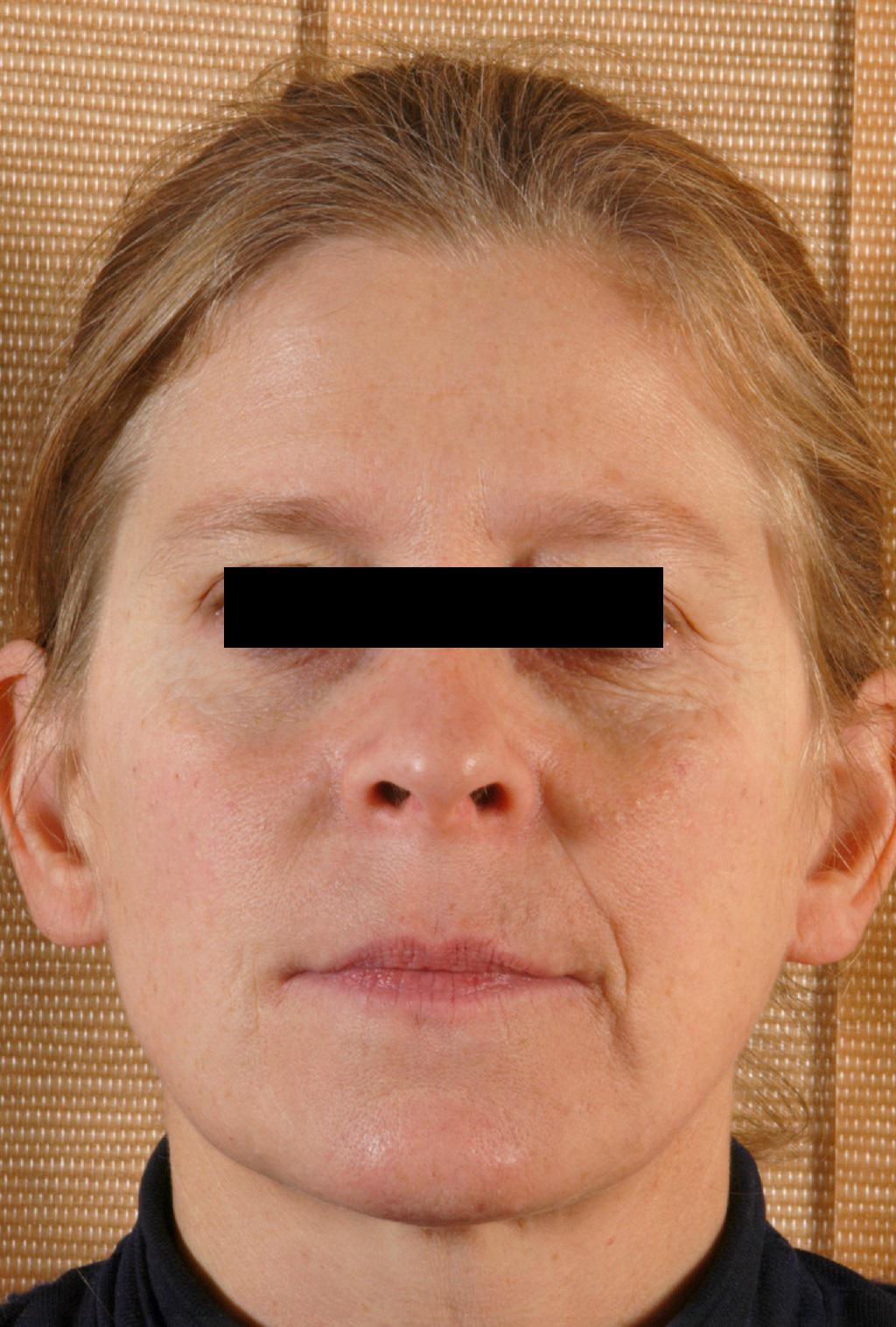}}\label{misc_side}}
\subfloat[]{\fbox{\includegraphics[height=2.5cm]{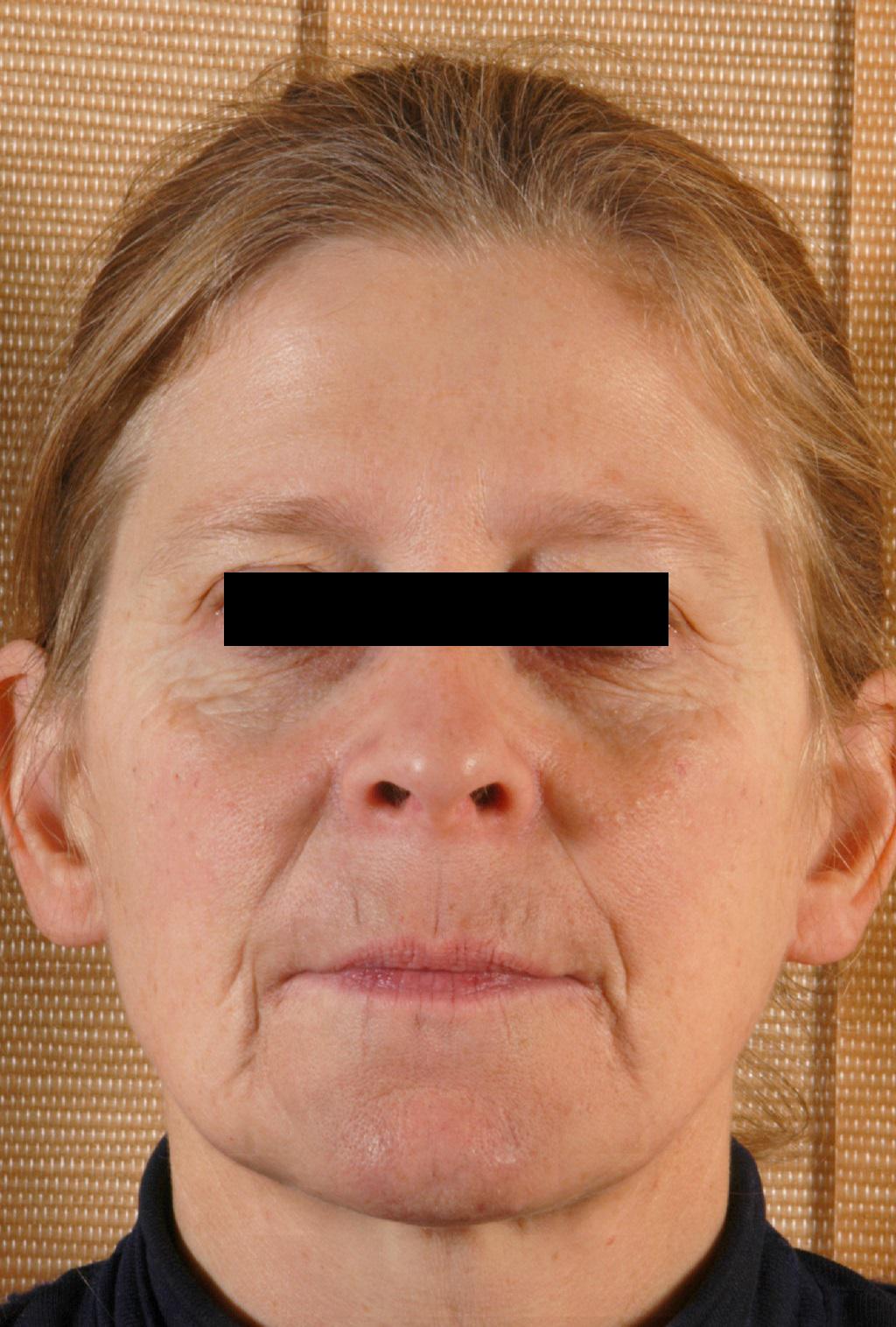}}\label{misc_bottom}}
\subfloat[]{\fbox{\includegraphics[height=2.5cm]{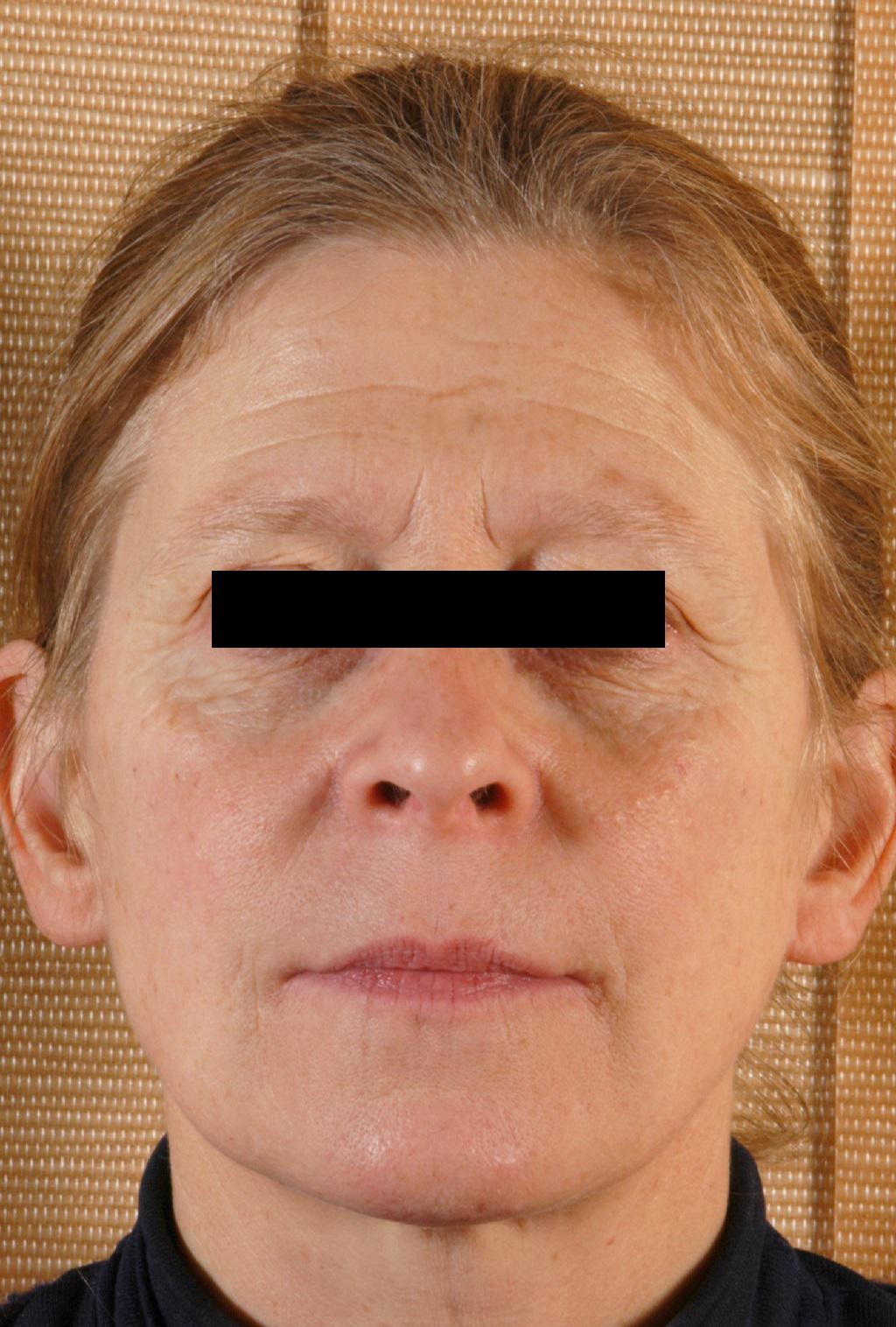}}\label{misc_top}}
\subfloat[]{\fbox{\includegraphics[height=2.5cm]{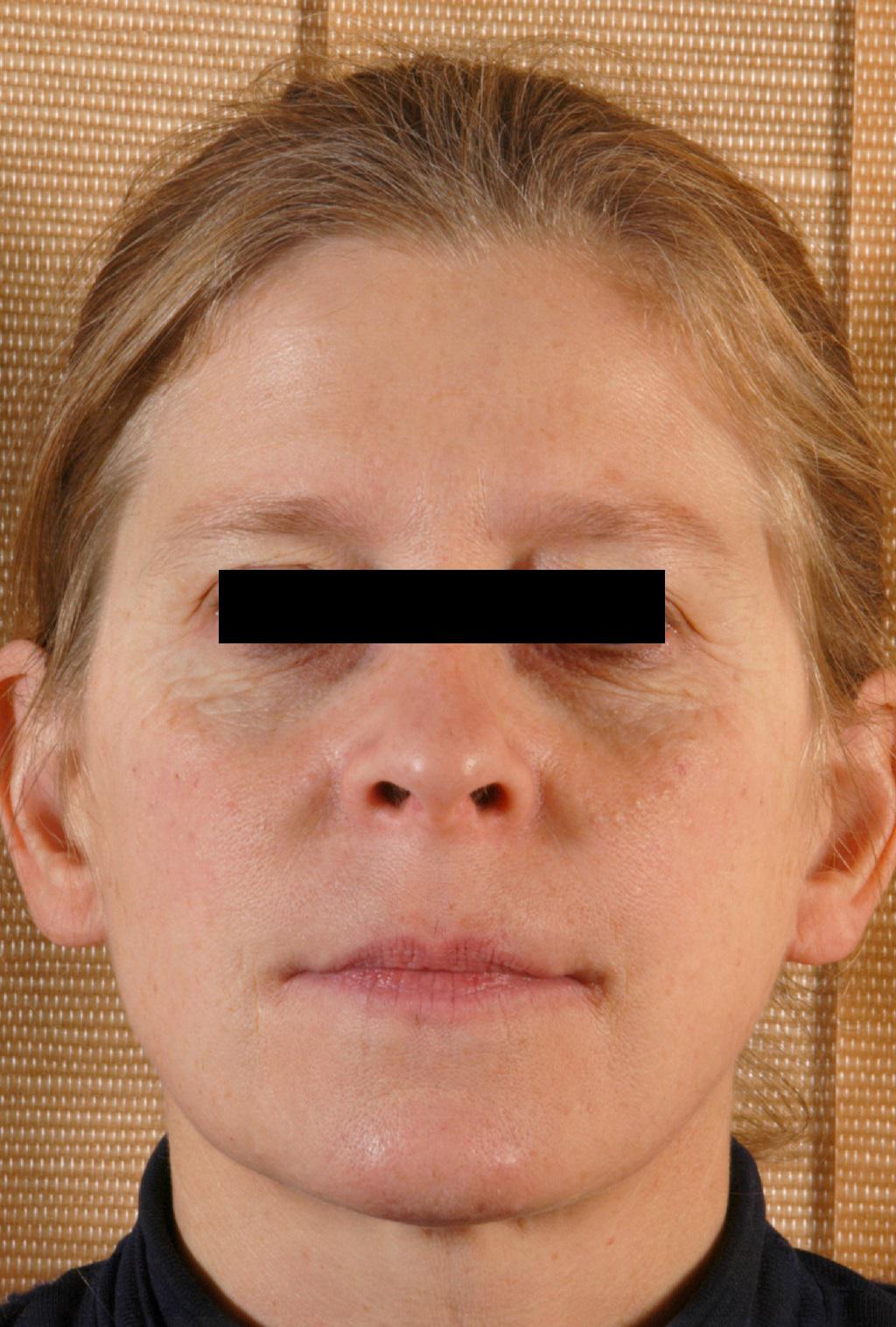}}\label{misc_underneath}}
\subfloat[]{\fbox{\includegraphics[height=2.5cm]{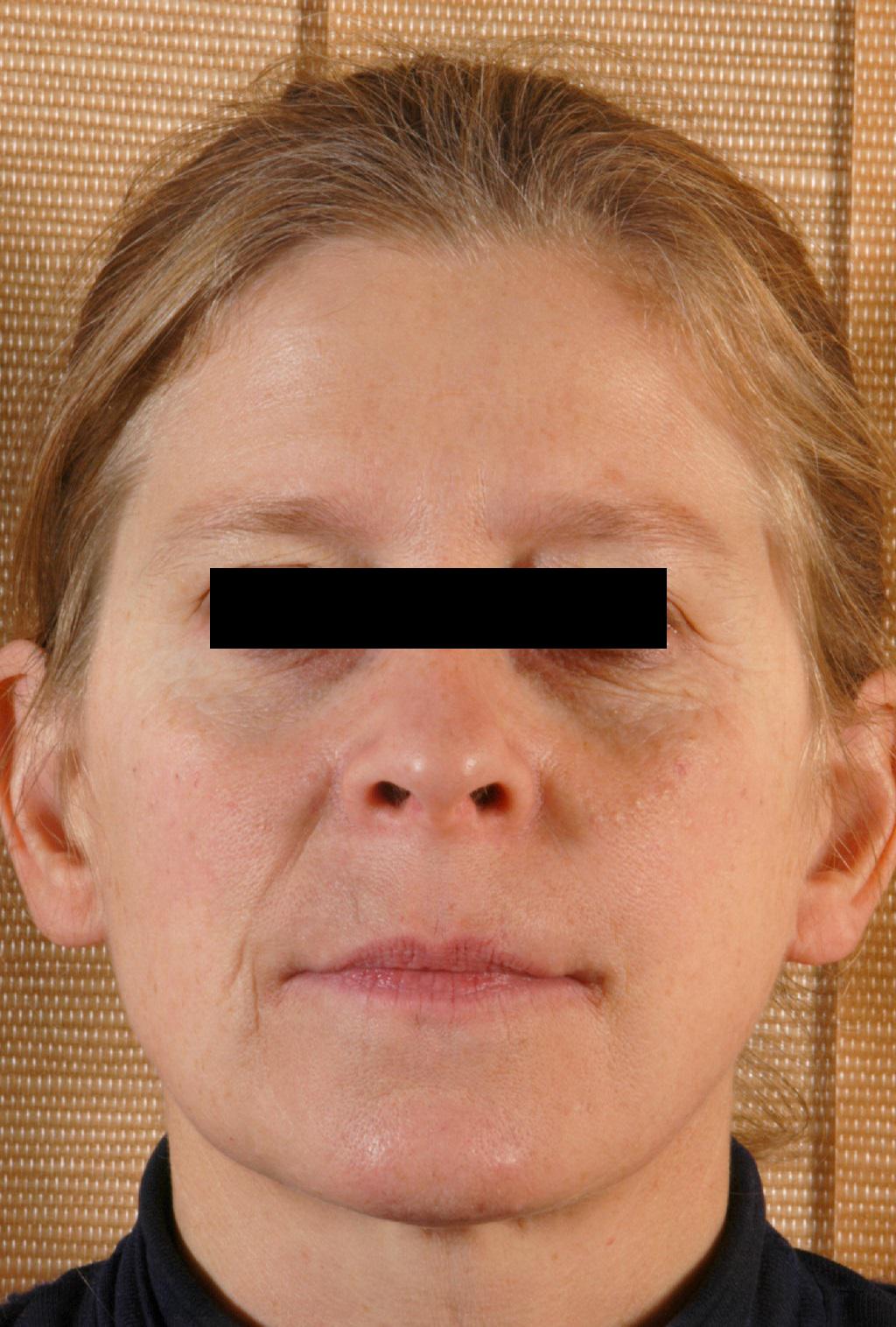}}\label{misc_asymmetric}}

\caption{Same face \protect\subref{misc_original} aged with different aging maps. \protect\subref{misc_side} rejuvenates all signs except for the nasolabial, corner of the lips and underneath the eyes wrinkles on the right part of the face. \protect\subref{misc_bottom} only ages the bottom of the face and \protect\subref{misc_top} only the top. \protect\subref{misc_underneath} only ages the wrinkles underneath the eye. \protect\subref{misc_asymmetric} ages the face in an asymmetric fashion, namely the right wrinkles underneath the eyes and the left nasolabial fold}
\label{fig:misc}
\end{figure}

\subsection{Evaluation Metrics}
To be considered successful, the task of face aging requires three criteria to be met: the image must be realistic, the identity of the subject must be preserved, and the face must be aged. These are respectively enforced during training thanks to the WGAN objective function, cycle-consistency loss, and aging map estimation loss. By nature, one single metric couldn't ensure that all criteria are met. For instance, the model could leave the input image without altering it, and still succeed in realism and identity. Contrarily, the model could succeed in aging but fail realism and/or identity. If one model isn't superior to another on every metric, we need to choose a trade-off. 

Our experiments on FFHQ and our high-quality standardized dataset never displayed any issue in the preservation of the subject identity. This is a consequence of the use of the attention mechanism that allows the generator to preserve the key facial features of the face. As a result, we chose to focus on the realism and aging criteria for our quantitative evaluation. Because our approach focuses on aging as a combination of aging signs instead of relying solely on age, we don't use the accuracy of the target age as a metric. Instead, we use the Fr\'echet Inception Distance (FID)\cite{heusel2017gans} to assess the realism of our images, and the Mean Average Error (MAE) for the accuracy of the target aging signs.

To do so, we use half of our dataset as a reference for real images, and the rest as the images to be transformed by our model. The aging maps used to transform these images are chosen randomly from the ground truth labels to ensure a distribution of generated images that follows the original dataset. We estimate the value of individual scores on all generated images using dedicated aging sign estimation models based on the ResNet\cite{szegedy2017inception} architecture. As a reference for the FID scores, we compute the FID between both halves of the real image dataset. Note that the size of our dataset prevents us from computing the FID on the recommended 50k+\cite{heusel2017gans, karras2019style}, thus leading to the overestimation of the value. This can be seen when computing the FID between real images only, giving a baseline FID of $49.0$. The results are presented in Table~\ref{table:metrics}.


\begin{table}
\setlength{\tabcolsep}{8pt}
    \begin{center}
        \caption{Fr\'echet Inception Distance and Mean Average Error for our different models}
        \label{table:metrics}
        \begin{tabular}{c|c|c|c}
            \hline
            {\bf Method} & {\bf Patch Size} & {\bf FID}$\downarrow$ & {\bf MAE}$\downarrow$\\
            \hline
            \hline
            Real Images & - & 49.0 & -\\
            \hline
            AMGAN (Ours) & $512 \times 512$ & {\bf 110.1} & {\bf 0.14}\\
            \hline
            AMGAN (Ours) & $256 \times 256$ & 110.7 & {\bf 0.14}\\
            w/o Aging Maps & $256 \times 256$ & 141.6 & 0.17\\
            \hline
            AMGAN (Ours) & $128 \times 128$ & 112.9 & 0.17\\
            w/o Location Maps & $128 \times 128$ & 140.0 & 0.20\\
            \hline
        \end{tabular}
    \end{center}
\end{table}

\subsection{Comparison Between Age and Clinical Signs}
When trained without clinical signs, using only the age to create a uniform aging map, the model still gives convincing results, with low FID and estimated age MAE. (Table~\ref{table:metrics_age}).

By comparing the results with the full aging map approach, however, it appears that some wrinkles don't exhibit their full range of dynamics. This is due to the fact that not all aging signs need to be maximized in order to reach the limit age of the dataset. In fact, the 150 oldest individuals of our standardized dataset (65 to 80 years old) display a median standard deviation of their normalized aging signs of $0.18$, highlighting the many possible combinations of aging signs in old people (Supplementary Materials, Fig. 1). For example, signs such as the forehead wrinkles are highly dependant on the facial expressions of the subject and are integral parts of the aging process. This an issue for the age-only model because it only offers one way to age a face.

To the contrary, the faces aged with the aging map offer much more control over the aging process. By controlling each individual sign of aging, we can choose whether to apply these expression wrinkles or not. A natural extension of this benefit is the pigmentation of the skin, which is viewed in some Asian countries as a sign of aging. An age-based model cannot produce aging for these countries without having to re-estimate the age from the local perspective. This doesn't scale, unlike our approach which, once trained with every relevant aging sign, can offer a face aging experience customized to the point of view of different countries, all in a single model and without additional labels.

\newcommand{\cmark}{\textcolor{Green}{\ding{51}}}
\newcommand{\xmark}{\textcolor{Red}{\ding{55}}}
\begin{table}
\setlength{\tabcolsep}{8pt}
    \begin{center}
        \caption{Fr\'echet Inception Distance and Mean Average Error for our model with clinical signs, and with age only}
        \label{table:metrics_age}
        \begin{tabular}{c|c|c|c|c}
            \hline
            {\bf Method} & {\bf Patch Size} & {\bf Control} & {\bf FID} & {\bf MAE}\\
            \hline
            \hline
            AMGAN (Ours) & $256 \times 256$ & \cmark & 110.7 & 0.143\\
            w/o Clinical Signs & $256 \times 256$ & \xmark & 101.3 & 0.116\\
            \hline
        \end{tabular}
    \end{center}
\end{table}

\subsection{Ablation Study}
\subsubsection{Effect of Patch Size}
\begin{figure}
    \centering
    \begin{minipage}{0.13\textwidth}
    Patch Size
    \end{minipage}
    \begin{minipage}{0.3\textwidth}
        \includegraphics[width=\textwidth]{eccv2020kit/imgs/other_papers/small_spacer.jpg}
    \end{minipage}
    \begin{minipage}{0.3\textwidth}
        \includegraphics[width=\textwidth]{eccv2020kit/imgs/other_papers/small_spacer.jpg}
    \end{minipage}
    
    \begin{minipage}{0.13\textwidth}
    $128 \times 128$
    \end{minipage}
    \begin{minipage}{0.3\textwidth}
        \fbox{\includegraphics[height=1.7cm]{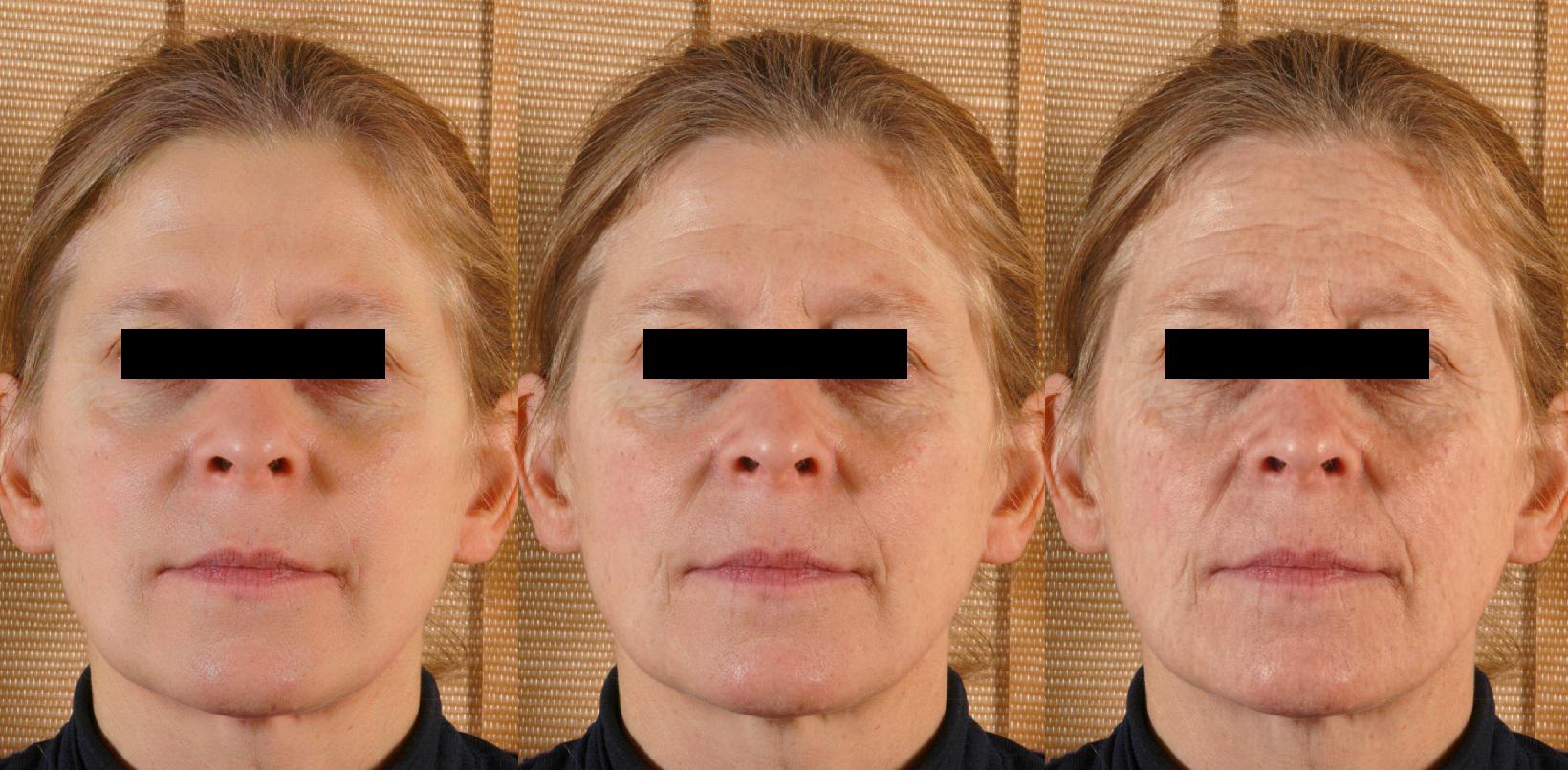}}
    \end{minipage}
    \begin{minipage}{0.3\textwidth}
        \fbox{\includegraphics[height=1.7cm]{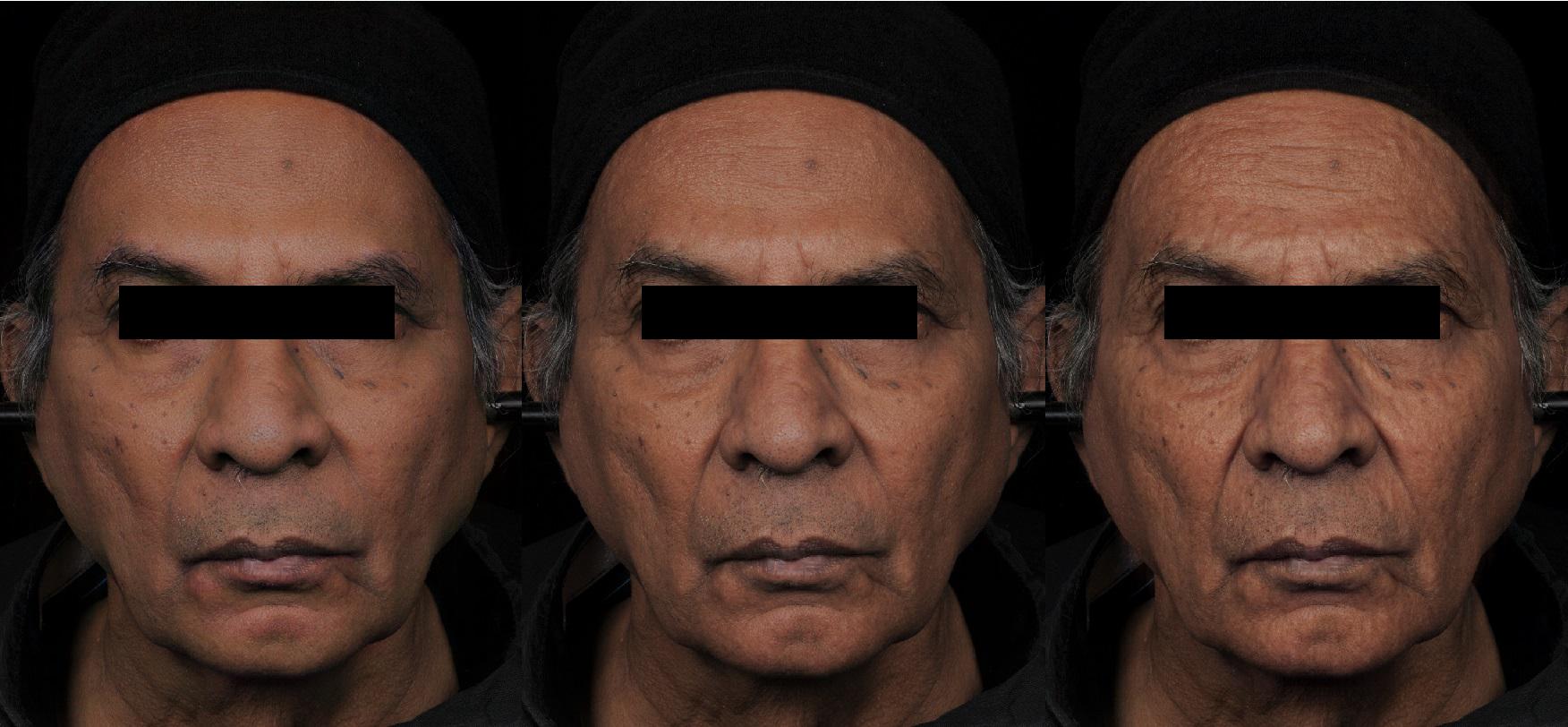}}
    \end{minipage}
    
    \begin{minipage}{0.13\textwidth}
    $256 \times 256$
    \end{minipage}
    \begin{minipage}{0.3\textwidth}
        \fbox{\includegraphics[height=1.7cm]{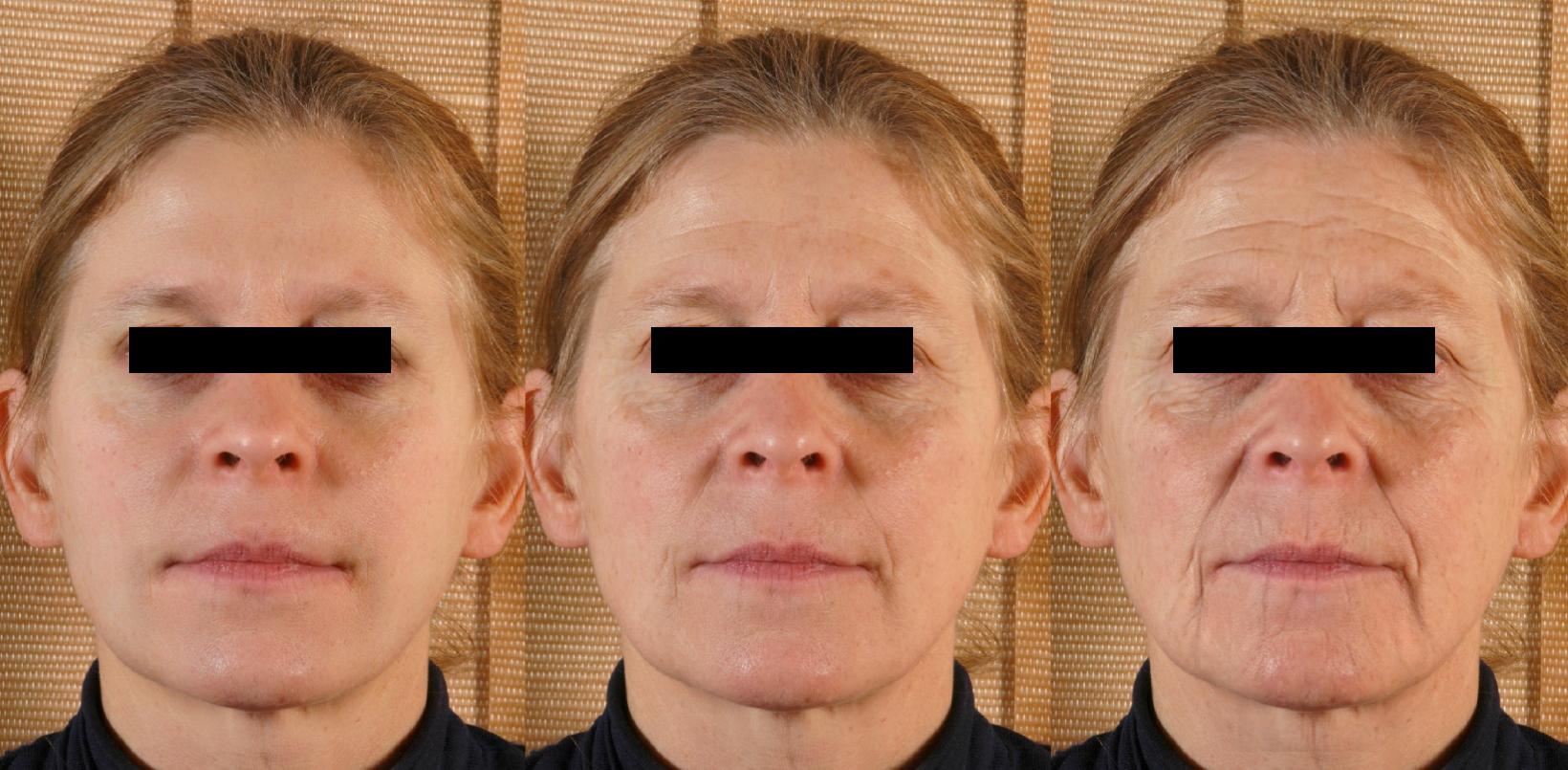}}
    \end{minipage}
    \begin{minipage}{0.3\textwidth}
        \fbox{\includegraphics[height=1.7cm]{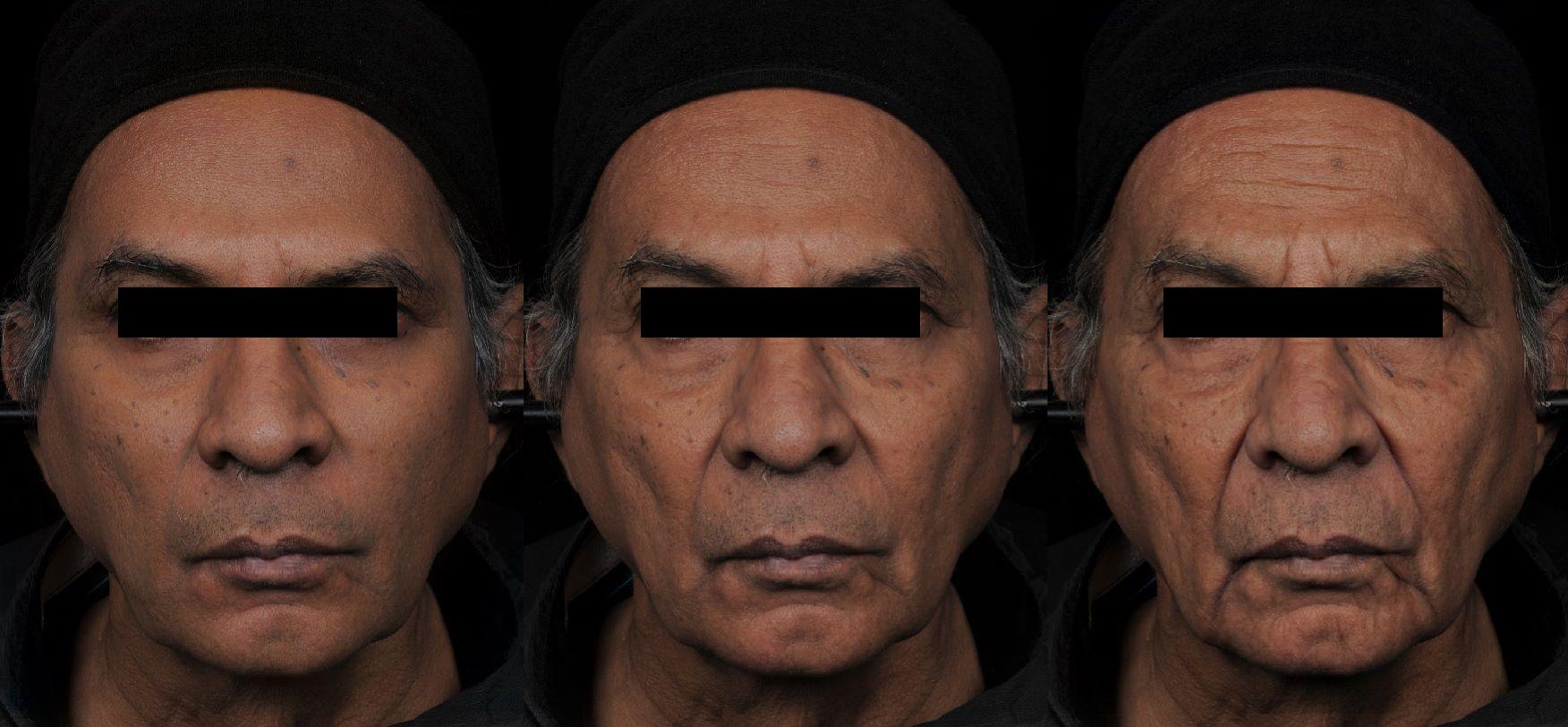}}
    \end{minipage}
    
    \begin{minipage}{0.13\textwidth}
    $512 \times 512$
    \end{minipage}
    \begin{minipage}{0.3\textwidth}
        \fbox{\includegraphics[height=1.7cm]{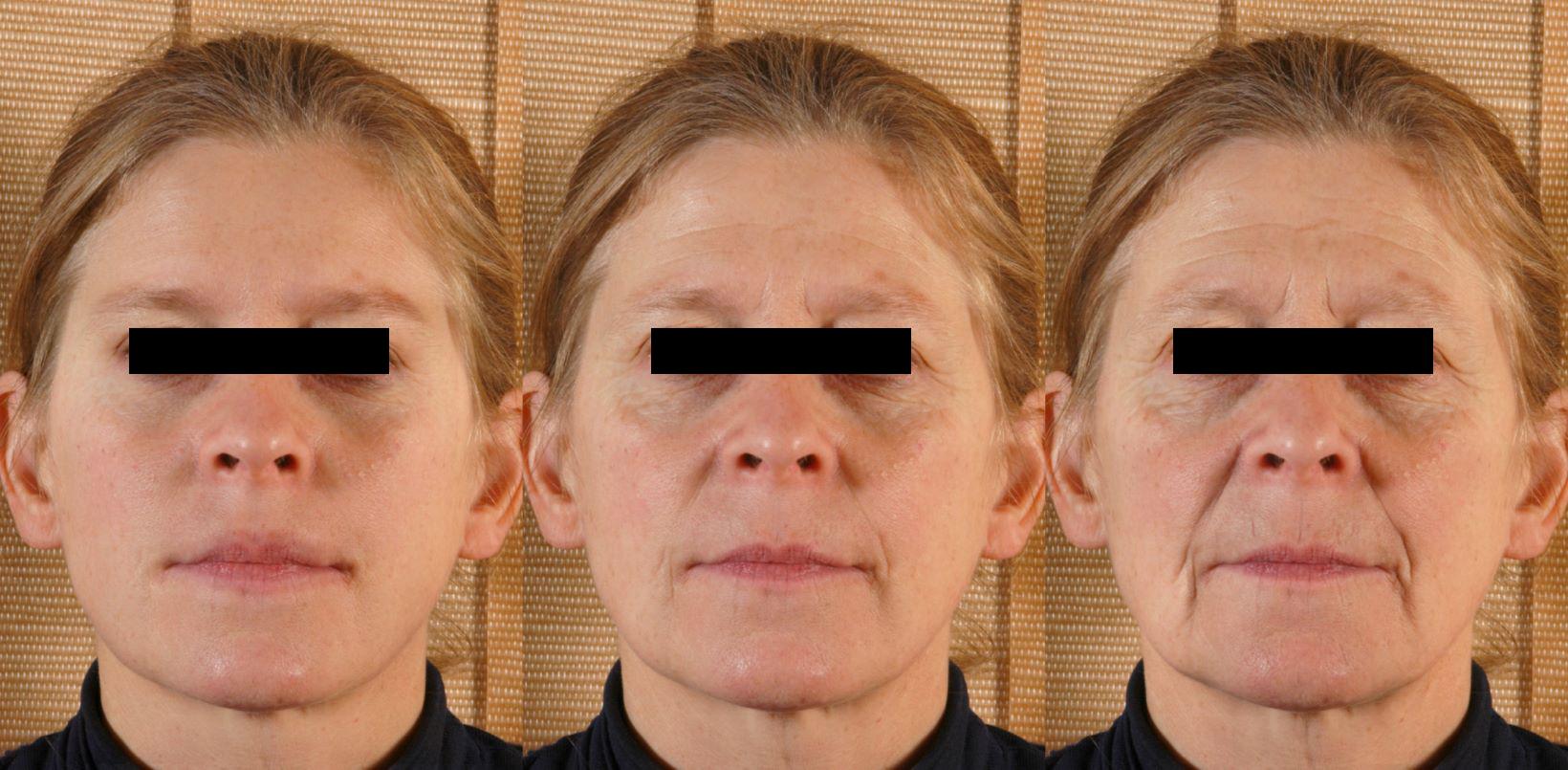}}
    \end{minipage}
    \begin{minipage}{0.3\textwidth}
        \fbox{\includegraphics[height=1.7cm]{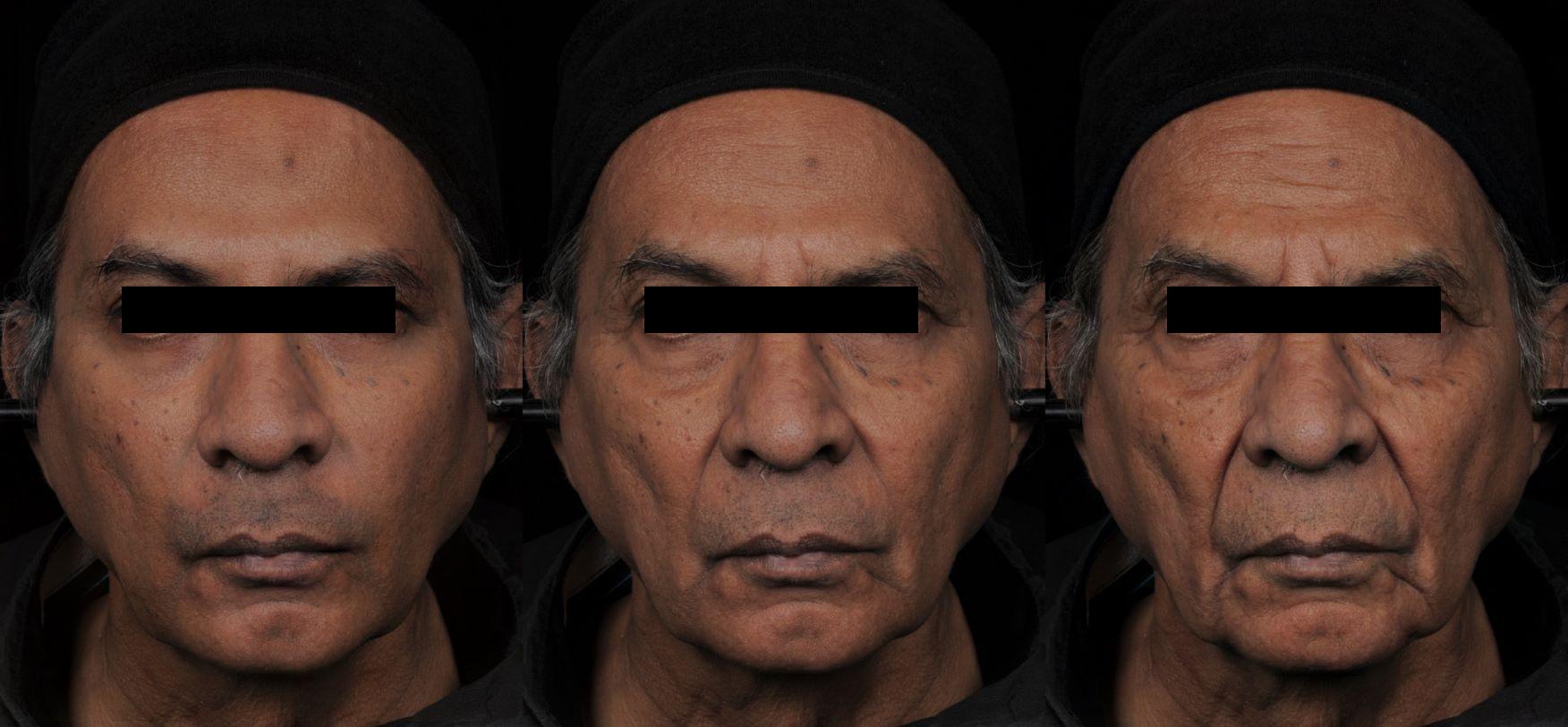}}
    \end{minipage}
    \caption{Results of rejuvenation ({\it left}) and aging ({\it right}) for different patch sizes on a $1024 \times 1024$ image}
    \label{fig:patch_size}
\end{figure}

When training the model for a given target image resolution ($1024 \times 1024$ pixels in our experiments), we can choose the size of the patch used for the training. The bigger the patch, the more context the model will have to perform the aging task. For the same computation power, however, larger patches cause the batch size to be smaller, which hinders the training\cite{brock2018large}. We conducted experiment using patches of $128 \times 128$, $256 \times 256$ and $512 \times 512$ pixels. Fig.~\ref{fig:patch_size} shows that all patch sizes manage to age the high-resolution face but to various degrees of realism. The smallest patch size suffers most from the lack of context and produces results that are inferior to the other two, with visible texture artifacts. The $256 \times 256$ patch gives convincing results, with minor imperfection only visible when compared to the $512 \times 512$ patch. These results suggest that we could apply this technique to larger resolutions, such as with patches of $512 \times 512$ on $2048 \times 2048$ images.

\subsubsection{Location Maps}
\begin{figure}
    \centering
    \fbox{\includegraphics[height=2cm]{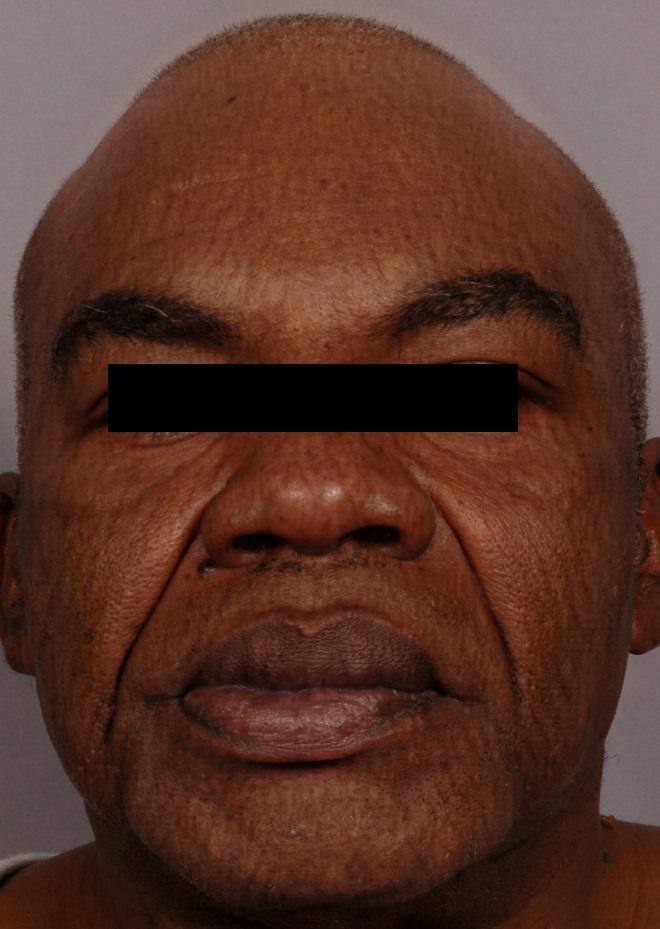}}
    \fbox{\includegraphics[height=2cm, width=2cm]{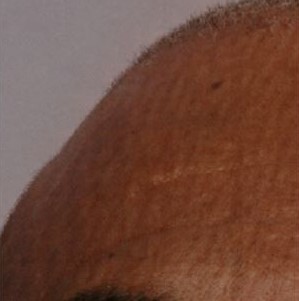}}
    \fbox{\includegraphics[height=2cm, width=2cm]{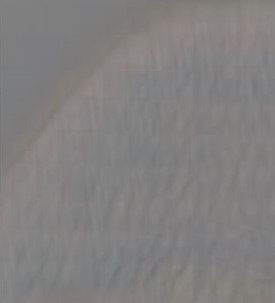}}
    \quad
    \fbox{\includegraphics[height=2cm]{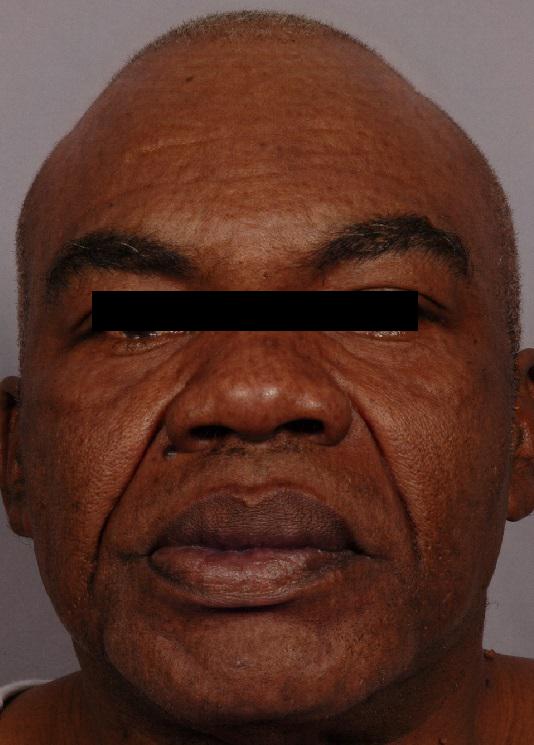}}
    \fbox{\includegraphics[height=2cm, width=2cm]{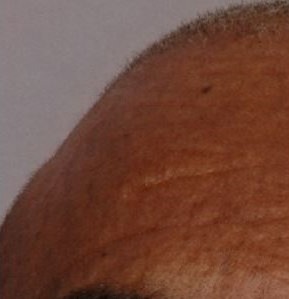}}
    \fbox{\includegraphics[height=2cm, width=2cm]{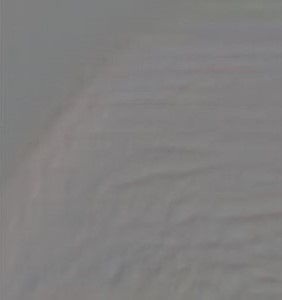}}
    \caption{Face aged with smallest patch size without ({\it left}) and with ({\it right}) location maps, along with the difference with the original image. The location maps eliminate the presence of diagonal texture artifacts, especially on the forehead where they allow horizontal wrinkle to appear}
    \label{fig:xy}
\end{figure}

To see the contribution of the location maps, we compared our model trained with and without them. As expected, the effect of the location maps is more prominent on small patch sizes, where the ambiguity is high. Fig.~\ref{fig:xy} shows how on small patch sizes and in the absence of location information, the model is unable to differentiate similar patches from different parts of the face. It is, therefore, unable to add wrinkles that are coherent with the location, and generates generic diagonal ripples. This effect is less present on larger patch sizes because the location of the patch is less ambiguous.

\subsubsection{Spatialization of Information}
\begin{figure}
    \begin{minipage}{\textwidth}
        \centering
        \fbox{\includegraphics[height=2cm]{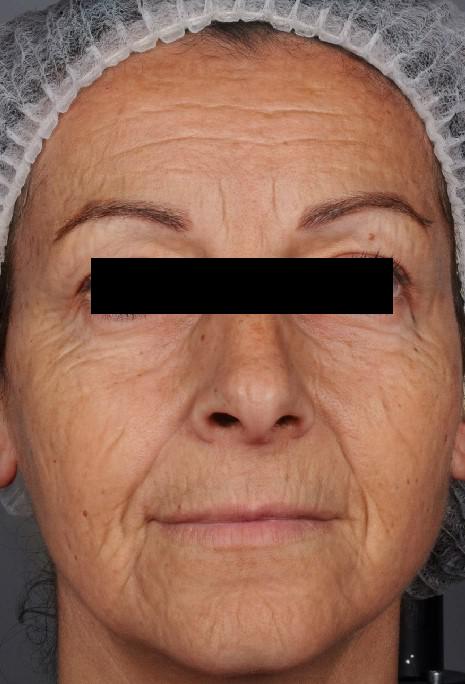}}
        \fbox{\includegraphics[width=2cm, height=2cm]{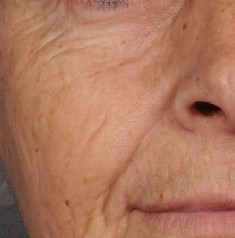}}
        \fbox{\includegraphics[width=2cm, height=2cm]{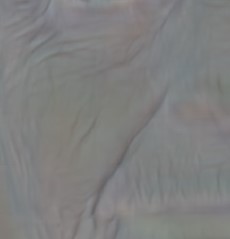}}
        \quad
        \fbox{\includegraphics[height=2cm]{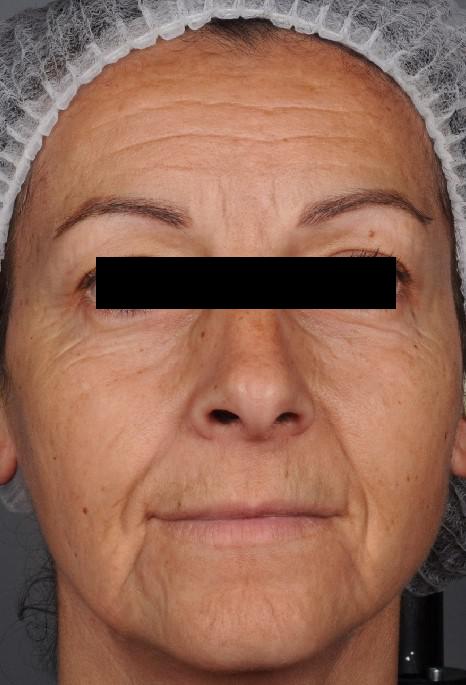}}
        \fbox{\includegraphics[width=2cm, height=2cm]{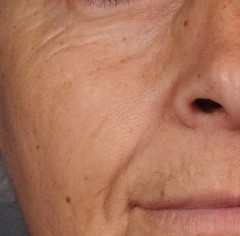}}
        \fbox{\includegraphics[width=2cm, height=2cm]{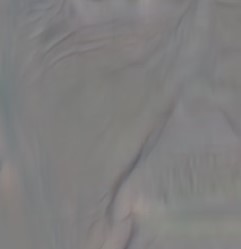}}
    \end{minipage}
    \caption{Face aged with medium patch size with individual uniform condition feature maps ({\it left}) and proposed aging maps ({\it right}), along with the difference with the original image. The aging maps help make the training more efficient thanks to denser spatialized information, and produce more realistic aging. The difference highlights the small unrealistic wrinkles for the baseline technique}
    \label{fig:multi}
\end{figure}

We compare our proposed aging maps against the baseline method of formatting conditions, namely to give all sign scores as individual uniform feature maps. Since not every sign is present in the patch, especially when the patch size is small, most of the processed information is of no use to the model. The aging maps represent a simple way of only giving the model the labels present in the patch, in addition to their spatial extent and location. Fig.~\ref{fig:multi} highlights the effect of the aging map. On small patches ($128 \times 128$, $256 \times 256$ pixels), the model struggles to create realistic results. The aging map helps reduce the complexity of the problem.

\section{Conclusion}
In this paper, we presented the use of clinical signs to create aging maps for face aging. Thanks to this technique, we demonstrated state-of-the-art results on high-resolution images with complete control over the aging process. Our patch-based approach allows conditional generative adversarial networks to be trained on large images while keeping a large batch size. This technique is applicable to various problems and can be used to tackle high-resolution problems with limited computational resources. In the future, the use of longitudinal data following the same person over time would allow a better understanding of the evolution of the aging signs on an individual basis, and therefore better personalizing of face aging factoring lifestyle, environmental and behavioral components.

\subsubsection{Acknowledgements}
We would like to thank Axel Sala-Martin for his insight on the model architecture and training process, and Robin Kips for many helpful discussions.

\clearpage
\bibliographystyle{splncs04}
\bibliography{egbib}
\end{document}